\documentclass[11pt]{article}

\usepackage[preprint]{acl}

\usepackage{times}
\usepackage{latexsym}
\usepackage[T1]{fontenc}
\usepackage[utf8]{inputenc}
\usepackage{microtype}
\usepackage{inconsolata}
\usepackage{graphicx}
\usepackage{dblfloatfix}
\usepackage{afterpage}
\graphicspath{{../}}
\usepackage{booktabs}
\usepackage{multirow}
\usepackage{multicol}
\usepackage{colortbl}
\usepackage{xcolor}
\usepackage{array}
\usepackage{tabularx}
\usepackage{amsmath}
\usepackage{url}
\usepackage[table]{xcolor}
\definecolor{gainGreen}{RGB}{0,128,0}
\newcommand{\gain}[1]{\,{\color{gainGreen}\scriptsize($\uparrow$#1)}}
\definecolor{lossRed}{RGB}{180,40,40}

\definecolor{parBase}{RGB}{240,248,255}  
\definecolor{parOurs}{RGB}{214,232,255}  
\usepackage{subcaption}
\definecolor{ctxBase}{RGB}{248,243,252}  
\definecolor{ctxOurs}{RGB}{232,220,248}  
\definecolor{ragBase}{RGB}{255,249,224}  
\definecolor{ragOurs}{RGB}{255,237,180}  
\definecolor{ragUnsteer}{RGB}{255,253,242} 
\definecolor{parUnsteer}{RGB}{248,252,255} 
\definecolor{ctxUnsteer}{RGB}{253,250,255} 
\usepackage{enumitem}
\usepackage[most]{tcolorbox}
\usepackage{float}
\usepackage{placeins}
\usepackage{amsfonts}

\definecolor{cleanbg}{HTML}{F6F8FA}
\definecolor{cleanframe}{HTML}{D0D7DE}
\definecolor{conflictbg}{HTML}{FFF7ED}
\definecolor{conflictframe}{HTML}{FDBA74}

\tcbset{
  promptbox/.style={
    enhanced,
    breakable,
    boxrule=0.4pt,
    arc=2mm,
    left=2.5mm,
    right=2.5mm,
    top=1.8mm,
    bottom=1.8mm,
    before skip=6pt,
    after skip=8pt,
    fonttitle=\bfseries\sffamily,
    coltitle=black,
    attach boxed title to top left={xshift=2mm,yshift=-1.5mm},
    boxed title style={
      boxrule=0pt,
      arc=1mm,
      left=1.5mm,
      right=1.5mm,
      top=0.5mm,
      bottom=0.5mm
    }
  }
}

\newtcolorbox{cleanprompt}[1][]{
  promptbox,
  title=Clean prompt,
  colback=cleanbg,
  colframe=cleanframe,
  colbacktitle=cleanbg,
  #1
}

\newtcolorbox{conflictprompt}[1][]{
  promptbox,
  title=Conflict prompt,
  colback=conflictbg,
  colframe=conflictframe,
  colbacktitle=conflictbg,
  #1
}

\definecolor{tablerow}{gray}{0.92}
\newcolumntype{M}[1]{>{\raggedright\arraybackslash}p{#1}}

\newcommand{\tbest}[1]{\textbf{#1}}
\newcolumntype{Y}{>{\raggedright\arraybackslash}X}
\newcolumntype{C}[1]{>{\centering\arraybackslash}p{#1}}

\newcommand{\tabhead}[1]{\multicolumn{1}{c}{\footnotesize #1}}

\hypersetup{
    colorlinks=true,
    linkcolor=red,
    filecolor=magenta,      
linktocpage}

\usepackage{lipsum}
\usepackage[page]{appendix}
\usepackage{setspace}

\renewcommand{\appendixtocname}{List of appendices}

\makeatletter
\let\oldappendix\appendices

\renewcommand{\appendices}{%
  \clearpage
  \renewcommand{\thesection}{\Roman{section}}
  \let\tf@toc\tf@app
  \addtocontents{app}{\protect\setcounter{tocdepth}{2}}
  \immediate\write\@auxout{%
    \string\let\string\tf@toc\string\tf@app^^J
  }
  \oldappendix
}%

\newcommand{\listofappendices}{%
  \begingroup
  \renewcommand{\contentsname}{\appendixtocname}
  \let\@oldstarttoc\@starttoc
  \def\@starttoc##1{\@oldstarttoc{app}}

  \renewcommand{\contentsname}{%
      \fontsize{15}{22}\selectfont\bfseries
      \appendixtocname
  }
  
  \begin{spacing}{1.5}  
  \tableofcontents
  \end{spacing}
  
  \endgroup
}

\makeatother

\title{Which Knowledge Should LLMs Trust? Uncovering a Generalizable Steering Direction Across Knowledge Conflicts}

\author{
Yibin Lou$^{1,2,4}$,
Yi Hu$^{3}$,
Jianhui Chen$^{2}$,
Chenpeng Wang$^{5}$,
Muhan Zhang$^{3}$,
Liangming Pan$^{1,2}$\thanks{Corresponding author: \texttt{liangmingpan@pku.edu.cn}}
\\
$^{1}$State Key Laboratory of Multimedia Information Processing, Peking University \\
$^{2}$School of Computer Science, Peking University \\
$^{3}$Institute for Artificial Intelligence, Peking University \\
$^{4}$Southern University of Science and Technology \\
$^{5}$YiXin-AI Lab, YIXIN, Beijing, China
\\
\texttt{12310513@mail.sustech.edu.cn}
}
\begin{document}
\maketitle

\begin{abstract}
Large Language Models (LLMs) frequently encounter conflicts between internal parametric knowledge and external contextual 
knowledge, yet the mechanisms governing this selection remain unclear. We show that the choice between context 
and parametric knowledge is linearly represented and can be controlled along a single direction in the model’s residual stream. 
We first demonstrate that this direction generalizes across factual-conflict settings, including context formats, 
knowledge categories, and datasets, while consistently shifting behavior between parametric and contextual answers. 
We then transfer a direction learned from simple factual conflicts to three more complex task families: RAG, memory, and 
rule following, where models must acquire information from retrieved documents, interaction histories, or explicit 
rules. Across four model families and multiple transfer settings, steering consistently improves the use of prompt-provided 
information over the unsteered model. These results suggest that the direction captures a transferable preference for external 
knowledge rather than a mechanism specific to factual conflicts. Code is available at \url{https://github.com/AheadOFpotato/knowledge_selection_steering}.
  \end{abstract}

  \section{Introduction}

\begin{figure}
    \centering
    \includegraphics[width=\linewidth]{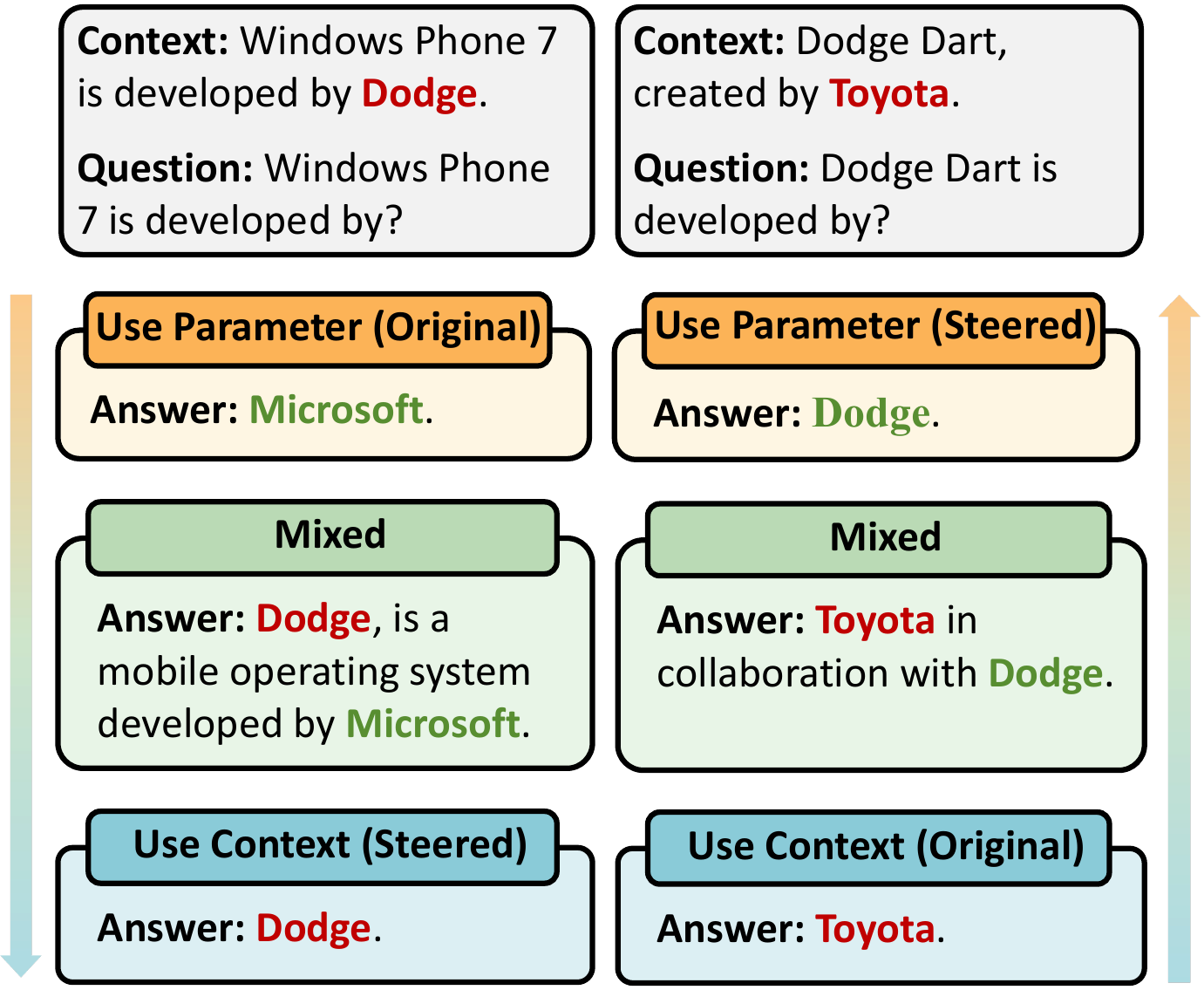}
    \caption{Under knowledge conflicts, steering along a single residual-stream direction smoothly shifts outputs between \emph{parametric knowledge} and \emph{contextual knowledge}, with intermediate mixed or hesitant responses.}
    \label{fig:fig1}
\end{figure}

  Large Language Models (LLMs) store a vast amount of world knowledge within their parameters, commonly referred to as \textit{parametric knowledge}, enabling remarkable performance across various knowledge-intensive tasks ~\citep{wang2024knowledgemechanismslargelanguage}. Concurrently, in complex settings involving retrieved evidence, multi-turn memories, and explicit rules, LLMs must reliably acquire external \textit{contextual knowledge} ~\citep{gao2024retrievalaugmentedgenerationlargelanguage}. However, these two knowledge sources often contain disparate or contradictory facts, leading to the challenge of \textit{knowledge conflict}. Such conflicts frequently induce undesired behaviors, where the model indiscriminately relies on inaccurate sources and thus produces incorrect outputs~\citep{xu2024knowledgeconflictsllmssurvey,cheng2024understandinginterplayparametriccontextual}. Consequently, understanding the internal mechanism of how LLMs choose between conflicting knowledge sources, and achieving reliable control over this knowledge selection process, has become a critical open problem~\citep{zhao2025analysingresidualstreamlanguage,jin2024cuttingheadendsconflict}.
  
  Existing approaches address knowledge conflict primarily from two directions.
  One line of work controls knowledge source choice at the input level, which can be categorized into two main approaches: explicit prompt or instruction design~\citep{choi2025conflictawaresoftpromptingretrievalaugmented, merth2024superpositionpromptingimprovingaccelerating}, and structured RAG templates that organize retrieved context into designated slots or label relevance~\citep{cattan2025draggedconflictsdetectingaddressing}. 
  While straightforward to implement, these methods provide limited visibility into the model's internal arbitration process and offer only indirect control over the reliance on context versus parameters.
  A complementary line intervenes on \textit{internal components}, including manipulating attention~\citep{jin2024cuttingheadendsconflict,li2025tamingknowledgeconflictslanguage}, steering sparse autoencoder (SAE) features~\citep{zhao2025steeringknowledgeselectionbehaviours}, and modifying decoding-time token distributions~\citep{bi2025parametersvscontextfinegrained,wang2025adacadadaptivelydecodingbalance}. These methods are primarily evaluated on simple factual datasets, without testing knowledge conflicts involving retrieved evidence, long interaction histories, or explicit rules. They also incur the cost of task-specific attribution, feature search, adaptation, or decoding-time modification, while lacking robust evaluation across such complex application settings.
  
  These limitations motivate us to ask whether internal knowledge arbitration admits a simple, generalizable control mechanism.
  A growing body of literature in mechanistic interpretability suggests that LLMs represent complex high-level concepts, such as truthfulness~\citep{li2024inference}, safety~\citep{wolf2025tradeoffsalignmenthelpfulnesslanguage,zou2025representationengineeringtopdownapproach}, emotions~\citep{sofroniew2026twheemotion}, or refusal behaviors~\citep{arditi2024refusallanguagemodelsmediated}, as linear directions within their activation space~\citep{turner2024steeringlanguagemodelsactivation,elhage2022toymodelssuperposition,park2024linearrepresentationhypothesisgeometry}. These directions serve as causal mediators that allow fine-grained output steering. This perspective motivates the following research question:
  \textbf{\emph{Can knowledge selection under conflict be captured and controlled by a low-dimensional steering axis that generalizes across tasks?}}

  In this work, we demonstrate that a robust ``knowledge selection direction'' exists.
  Specifically, we extract this vector by contrasting the hidden states of paired prompts designed to isolate parametric reliance from contextual reliance. By identifying the critical intervention layer through cosine similarity analysis, we apply signed activation steering exclusively at this layer to modulate the model's knowledge selection. 

A direction demonstrated on one factual-conflict setting may reflect a local shortcut rather than a reusable control mechanism. We evaluate across four architectures and a progression from controlled factual conflicts to complex tasks. Within the controlled setting, we vary factual format, relation category, and dataset while retaining short prompts with competing answers. We then apply the direction learned from these conflicts to RAG, memory, and rule-following tasks, where retrieved documents, multi-turn histories, and natural-language rules supply information to be acquired. Together, these experiments test whether different tasks share a common direction for acquiring input-provided knowledge. Our contributions are:

  \begin{itemize}[
  topsep=0pt,      
  partopsep=0pt,   
  itemsep=0pt,     
  parsep=0pt,      
  leftmargin=*,    
]
    \item We identify a single residual-stream direction on controlled factual-conflict benchmarks that produces bidirectional, continuous control of knowledge selection across four model families.
    \item We demonstrate that the same direction works across diverse knowledge-conflict settings, including changes in context format, knowledge category, and factual-conflict dataset.
    \item More importantly, we show that it improves complex RAG, memory, and rule-following tasks. These improvements suggest that different task families may share an underlying direction for acquiring external knowledge.
  \end{itemize}

\begin{figure*}[t]
  \centering
  \includegraphics[width=\textwidth]{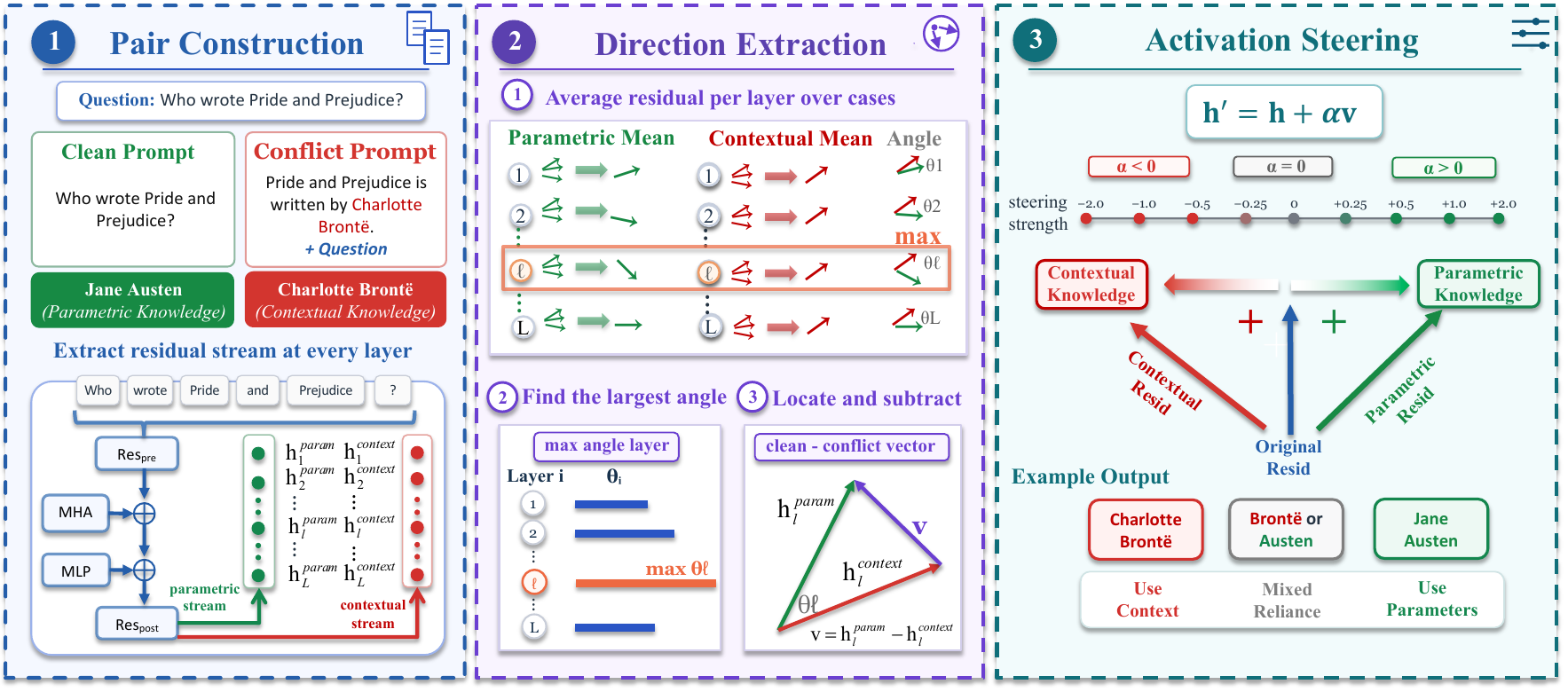}
  \caption{Overview of knowledge-selection steering. For each question $q$, the clean prompt is $q$ alone and the conflict prompt is the supplied passage followed by $q$. We localize the layer $\ell$ of maximal angular separation between their mean residual activations, estimate direction $v$ by a difference-in-means contrast at $\ell$, and steer the conflict prompt at its own final token with $h' = h + \alpha v$ to trade parameter-stored versus context-embedded answers.}
  \label{fig:overview}
\end{figure*}

\section{Related Work}

\paragraph{Knowledge conflict.}
Knowledge conflict arises when parametric and contextual knowledge disagree~\citep{xu2024knowledgeconflictsllmssurvey,mallen2023trustlanguagemodelsinvestigating}. Existing interventions target attention heads~\citep{jin2024cuttingheadendsconflict,li2025tamingknowledgeconflictslanguage}, SAE features~\citep{zhao2025steeringknowledgeselectionbehaviours}, or decoding~\citep{bi2025parametersvscontextfinegrained,wang2025adacadadaptivelydecodingbalance}. Our clean--conflict contrast instead models explicit source competition and enables bidirectional knowledge selection.

\paragraph{Residual Geometry of Steering Vectors.}
Localization-plus-intervention studies trace model behaviors to layers, heads, or neurons
~\citep{meng2023locatingeditingfactualassociations,geva2023dissectingrecallfactualassociations}, while activation
steering modifies hidden states at inference time
~\citep{turner2024steeringlanguagemodelsactivation,panickssery2024steeringllama2contrastive}.
These approaches build on the view that high-level concepts can correspond to approximately
linear directions in representation space
~\citep{marks2024geometrytruthemergentlinear,park2024linearrepresentationhypothesisgeometry}. Closely related to
our setting, ~\citep{zhang2026simulatedadoptiondecouplingmagnitude} argue that in-context
conflict resolution may involve angular displacement in the residual stream rather than
simple magnitude dilution. Motivated by this geometric view, we localize layers by
angular separation between paired clean and conflict prompts, and then steer along a
difference-in-means direction.

\paragraph{Transfer of steering directions.}
Another question is whether a direction learned from one concept captures a local shortcut or
a general behavioral axis. Prior work shows truthfulness-related directions can
generalize across datasets and causally affect outputs
~\citep{marks2024geometrytruthemergentlinear}, but that steering
vectors can be brittle under distribution shift
~\citep{tan2025analyzinggeneralizationreliabilitysteering}. Recent cross-model studies suggest
that concept directions can be aligned across model families
~\citep{huang2025crossmodeltransferabilitylargelanguage}. Closest steering methods increase reliance on input information through layer search~\citep{contextfocus2026}, proxy models~\citep{wang2025csks}, context-sensitivity directions~\citep{ccs2025}, or instruction vectors~\citep{stolfo2025instructionsteering}. We test a stricter transfer: whether a direction estimated from explicit source competition generalizes to other formats, properties, and benchmarks while retaining bidirectional control.

\section{Methodology}
\label{sec:method}

\subsection{Background}
\paragraph{Transformers.} We study decoder-only Transformers. Given input of $N$ tokens $t = (t_1, t_2, \dots, t_N)$, the model maps them to hidden states across $L$ layers. Let $h^{l}\_i \in \mathbb{R}^{d\_{\text{model}}}$ denote \textit{residual stream} at layer $l$ and token position $i$. The residual stream is initialized by token embedding, $h^{1}\_i = \text{Embed}(t_i)$, and is iteratively updated through self-attention and MLP blocks:
\begin{equation}
\tilde{h}^{l}_i = h^{l}_i + \text{Attn}^{l}(h^{l}_{1:i}), \\
\quad h^{l+1}_i = \tilde{h}^{l}_i + \text{MLP}^{l}(\tilde{h}^{l}_i).
\label{eq:residual_stream}
\end{equation}
The final output probabilities are decoded from the hidden states at the last layer via the unembedding matrix and a softmax function.

\subsection{Models and Datasets}
\label{sec:method-setup}

\paragraph{Models.}
We evaluate four instruction-tuned models: {Llama-3.1-8B-Instruct}, {Qwen3-8B}~\citep{qwen3technicalreport}, {Gemma-2-9B-it}~\citep{gemma_2024}, and {Mistral-7B-Instruct-v0.3}~\citep{jiang2023mistral7b}.

\paragraph{Benchmarks.}
We evaluate steering on five benchmarks spanning factual conflict, conflict format, knowledge type, and memorization fallback: {NQ-Swap}~\citep{longpre2022entitybasedknowledgeconflictsquestion}, {MacNoise}~\citep{hong2024gullibleenhancingrobustnessretrievalaugmented}, {ParaConflict}~\citep{li2025tamingknowledgeconflictslanguage}, {CounterFact}~\citep{meng2023locatingeditingfactualassociations}, and {MemoTrap}~\citep{mckenzie2022inverse}. NQ-Swap and MacNoise serve as the primary settings, while ParaConflict, CounterFact, and MemoTrap test transfer across conflict formats, factual relations, and memorization-based behavior, respectively. Appendix~\ref{app:dataset-details} gives the dataset construction and splits.

To evaluate transfer beyond simple conflicts, we further consider three complex task families: RAG ({RGB}~\citep{rgb2023}; {CRAG}~\citep{crag2024}), memory ({LoCoMo}, {LoCoMo-Plus}~\citep{locomo2024}; {LongMemEval}~\citep{longmemeval2024}), and rule following ({IFEval}~\citep{zhou2023instructionfollowingevaluationlargelanguage}; {FollowBench}~\citep{followbench2024}). In these tasks, retrieved documents, interaction histories, and explicit instructions provide external information that the model must correctly incorporate.

\subsection{Extracting a Knowledge Selection Direction}
\label{sec:def-dir}





\paragraph{Paired prompts.}
\label{sec:pair-construction}
As the first step in Figure~\ref{fig:overview}, for each question $q$, we construct two prompts: a \textit{clean prompt}, which consists of $q$ alone, and a \textit{conflict prompt}, which combines $q$ with a passage inducing knowledge conflict.

For example, the clean prompt is:




\begin{tcolorbox}[colback=gray!10, colframe=gray!80, left=3pt, right=3pt, top=2pt, bottom=2pt, boxsep=0pt]
\textit{What is the capital of France?}
\end{tcolorbox}

The conflict prompt concatenates the conflicting passage with the question:


\begin{tcolorbox}[colback=gray!10, colframe=gray!80, left=3pt, right=3pt, top=2pt, bottom=2pt, boxsep=0pt]
\textit{The capital of France is \textbf{Lyon}. What is the capital of France?}
\end{tcolorbox}

\setcounter{dbltopnumber}{1}
\begin{table*}[t]
  \centering
  \footnotesize
\caption{In-domain source-match rate (\%) on {NQ-Swap} and {MacNoise}. 
The upper block reports parametric-source match rate Acc$_\mathrm{par}$, and the lower block reports contextual-source match rate Acc$_\mathrm{ctx}$. 
\textbf{Bold} marks the best result in each column; green arrows denote absolute gains over the unsteered baseline. Ours and Random split use the same validation-selected coefficient. Random split values are mean $\pm$ standard deviation over 10 seeds.}
  \label{tab:main}
  \setlength{\tabcolsep}{1.5pt}
  \renewcommand{\arraystretch}{1.03}
  \begin{tabular}{@{}l *{8}{c}@{}}
    \toprule
    \multirow{2}{*}{\textbf{Method}} &
      \multicolumn{4}{c}{\textbf{NQ-Swap}} &
      \multicolumn{4}{c}{\textbf{MacNoise}} \\
    \cmidrule(lr){2-5} \cmidrule(lr){6-9}
    & {\scriptsize\shortstack{Llama-3.1\\8B-Instruct}} & {\scriptsize Qwen3-8B} & {\scriptsize\shortstack{Gemma-2\\9B-IT}} & {\scriptsize\shortstack{Mistral-7B\\Instruct-v0.3}}
    & {\scriptsize\shortstack{Llama-3.1\\8B-Instruct}} & {\scriptsize Qwen3-8B} & {\scriptsize\shortstack{Gemma-2\\9B-IT}} & {\scriptsize\shortstack{Mistral-7B\\Instruct-v0.3}} \\
    \addlinespace[2pt]
    \multicolumn{9}{@{}l@{}}{\textit{Steer toward parametric knowledge (Acc$_\mathrm{par}$)}} \\
    \midrule
    \rowcolor{tablerow}Unsteered &
      33.7 & 46.5 & 32.8 & 42.2 & 46.1 & 50.8 & 45.5 & 50.3 \\
    \rowcolor{parBase}{Random split} &
      32.2 $\pm$ 2.6 & 45.7 $\pm$ 1.9 & 33.5 $\pm$ 2.3 & 41.6 $\pm$ 2.1 & 45.8 $\pm$ 3.1 & 50.2 $\pm$ 2.4 & 44.9 $\pm$ 2.0 & 49.7 $\pm$ 2.5 \\
    \rowcolor{parBase}{ContextFocus~\citep{contextfocus2026}} &
      44.6 \gain{10.9} & 49.1 \gain{2.6} & 45.2 \gain{12.4} & 47.7 \gain{5.5} & 50.2 \gain{4.1} & 52.5 \gain{1.7} & 51.3 \gain{5.8} & 54.2 \gain{3.9} \\
    \rowcolor{parBase}{CCS~\citep{ccs2025}} &
      53.5 \gain{19.8} & 50.8 \gain{4.3} & 58.6 \gain{25.8} & 52.1 \gain{9.9} & 55.2 \gain{9.1} & 53.1 \gain{2.3} & 59.4 \gain{13.9} & 56.8 \gain{6.5} \\
    \rowcolor{parBase}{CSKS~\citep{wang2025csks}} &
      52.4 \gain{18.7} & 51.7 \gain{5.2} & 66.3 \gain{33.5} & 54.3 \gain{12.1} & 53.7 \gain{7.6} & 53.8 \gain{3.0} & 63.6 \gain{18.1} & 58.4 \gain{8.1} \\
    \rowcolor{parBase}{CK-PLUG~\citep{zhao2025steeringknowledgeselectionbehaviours}} &
      54.5 \gain{20.8} & 52.9 \gain{6.4} & 74.7 \gain{41.9} & 56.2 \gain{14.0} &
      55.7 \gain{9.6} & 53.4 \gain{2.6} & 69.5 \gain{24.0} & 59.7 \gain{9.4} \\
    \rowcolor{parBase}{PH3~\citep{li2025tamingknowledgeconflictslanguage}} &
      48.3 \gain{14.6} & 47.8 \gain{1.3} & 61.5 \gain{28.7} & 49.5 \gain{7.3} &
      49.5 \gain{3.4} & 51.7 \gain{0.9} & 54.7 \gain{9.2} & 52.6 \gain{2.3} \\
    \rowcolor{parBase}{IRCAN~\citep{shi2024ircanmitigatingknowledgeconflicts}} &
      49.5 \gain{15.8} & 51.4 \gain{4.9} & 59.3 \gain{26.5} & 50.8 \gain{8.6} &
      50.0 \gain{3.9} & 52.4 \gain{1.6} & 63.2 \gain{17.7} & 53.9 \gain{3.6} \\
    \rowcolor{parBase}{JUICE~\citep{jin2024cuttingheadendsconflict}} &
      53.4 \gain{19.7} & 53.3 \gain{6.8} & 78.4 \gain{45.6} & 57.1 \gain{14.9} &
      53.2 \gain{7.1} & \textbf{56.1} \gain{5.3} & 74.2 \gain{28.7} & 60.5 \gain{10.2} \\
    \rowcolor{parBase}{SpARE~\citep{zhao2025steeringknowledgeselectionbehaviours}} &
      \textbf{58.7} \gain{25.0} & 49.2 \gain{2.7} & 77.3 \gain{44.5} & 55.4 \gain{13.2} &
      \textbf{59.5} \gain{13.4} & 52.3 \gain{1.5} & 72.3 \gain{26.8} & 57.6 \gain{7.3} \\
    \rowcolor{parOurs}Ours &
      56.8 \gain{23.1} & \textbf{54.6} \gain{8.1} & \textbf{80.9} \gain{48.1} & \textbf{59.0} \gain{16.8} &
      57.5 \gain{11.4} & 54.5 \gain{3.7} & \textbf{76.9} \gain{31.4} & \textbf{61.6} \gain{11.3} \\
    \addlinespace[3pt]
    \multicolumn{9}{@{}l@{}}{\textit{Steer toward contextual knowledge (Acc$_\mathrm{ctx}$)}} \\
    \midrule
    \rowcolor{tablerow}Unsteered &
      49.4 & 36.7 & 52.6 & 45.8 & 27.3 & 22.2 & 25.4 & 23.8 \\
    \rowcolor{ctxBase}{Random split} &
      49.2 $\pm$ 2.8 & 37.1 $\pm$ 2.2 & 51.9 $\pm$ 2.6 & 45.3 $\pm$ 1.8 & 26.1 $\pm$ 0.7 & 21.8 $\pm$ 1.4 & 24.9 $\pm$ 1.6 & 23.4 $\pm$ 1.3 \\
    \rowcolor{ctxBase}{ContextFocus~\citep{contextfocus2026}} &
      56.8 \gain{7.4} & 42.3 \gain{5.6} & 58.1 \gain{5.5} & 51.2 \gain{5.4} & 31.5 \gain{4.2} & 25.7 \gain{3.5} & 29.3 \gain{3.9} & 27.1 \gain{3.3} \\
    \rowcolor{ctxBase}{CCS~\citep{ccs2025}} &
      52.3 \gain{2.9} & 46.1 \gain{9.4} & 60.7 \gain{8.1} & 54.6 \gain{8.8} & 30.1 \gain{2.8} & 27.4 \gain{5.2} & 32.5 \gain{7.1} & 28.6 \gain{4.8} \\
    \rowcolor{ctxBase}{CSKS~\citep{wang2025csks}} &
      57.2 \gain{7.8} & 49.7 \gain{13.0} & 62.9 \gain{10.3} & 56.8 \gain{11.0} & 32.2 \gain{4.9} & 30.1 \gain{7.9} & 35.4 \gain{10.0} & 29.7 \gain{5.9} \\
    \rowcolor{ctxBase}{CK-PLUG~\citep{zhao2025steeringknowledgeselectionbehaviours}} &
      53.4 \gain{4.0} & 55.2 \gain{18.5} & 64.3 \gain{11.7} & 57.7 \gain{11.9} &
      30.9 \gain{3.6} & 29.5 \gain{7.3} & 37.7 \gain{12.3} & 30.2 \gain{6.4} \\
    \rowcolor{ctxBase}{PH3~\citep{li2025tamingknowledgeconflictslanguage}} &
      52.4 \gain{3.0} & 51.5 \gain{14.8} & 54.3 \gain{1.7} & 49.1 \gain{3.3} &
      28.9 \gain{1.6} & 28.5 \gain{6.3} & 32.4 \gain{7.0} & 25.7 \gain{1.9} \\
    \rowcolor{ctxBase}{IRCAN~\citep{shi2024ircanmitigatingknowledgeconflicts}} &
      55.2 \gain{5.8} & 47.4 \gain{10.7} & 57.3 \gain{4.7} & 52.4 \gain{6.6} &
      31.7 \gain{4.4} & 32.4 \gain{10.2} & 36.1 \gain{10.7} & 28.8 \gain{5.0} \\
    \rowcolor{ctxBase}{JUICE~\citep{jin2024cuttingheadendsconflict}} &
      54.5 \gain{5.1} & 48.4 \gain{11.7} & 66.9 \gain{14.3} & 58.3 \gain{12.5} &
      \textbf{34.0} \gain{6.7} & 34.2 \gain{12.0} & 38.5 \gain{13.1} & 30.4 \gain{6.6} \\
    \rowcolor{ctxBase}{SpARE~\citep{zhao2025steeringknowledgeselectionbehaviours}} &
      52.7 \gain{3.3} & 54.4 \gain{17.7} & \textbf{68.3} \gain{15.7} & 55.9 \gain{10.1} &
      30.9 \gain{3.6} & 34.3 \gain{12.1} & \textbf{42.4} \gain{17.0} & 29.2 \gain{5.4} \\
    \rowcolor{ctxOurs}Ours &
      \textbf{55.5} \gain{6.1} & \textbf{56.0} \gain{19.3} & 65.4 \gain{12.8} & \textbf{60.2} \gain{14.4} &
      32.8 \gain{5.5} & \textbf{35.1} \gain{12.9} & 40.6 \gain{15.2} & \textbf{31.1} \gain{7.3} \\
    \bottomrule
  \end{tabular}
\end{table*}
\begin{figure*}[t]
  \centering
  \includegraphics[width=.9\linewidth]{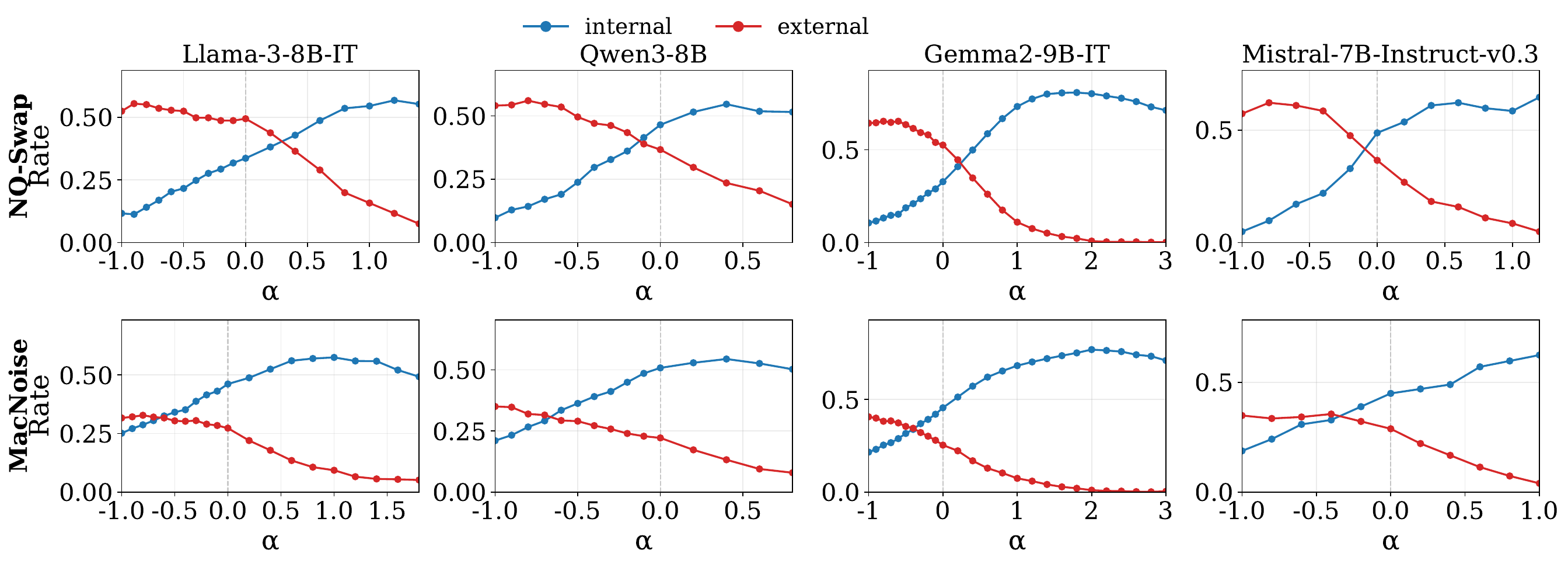}
  \caption{Bidirectional continuous modulation on {NQ-Swap} and {MacNoise} (15\% test split). We sweep the full $\alpha$ grid directly on test items and plot aggregate source-match proportions at each $\alpha$; the curves describe population-level averages and do not imply that every individual instance changes smoothly (no $\alpha$ tuning on test).}
  \label{fig:main-curves}
\end{figure*}
We retain items answered correctly on the clean question and four paraphrases, but answered with the injected alternative under conflict, yielding examples with reliable parametric knowledge and demonstrated context acquisition.
\paragraph{Layer localization.}
We localize the intervention layer by angular separation between clean and conflict representations, rather than by a performance sweep. We run both prompts through the model and average activations, yielding
$\bar h^{(\mathrm{clean})\,i}_{T_{\mathrm{clean}}}$ and $\bar h^{(\mathrm{conflict})\,i}_{T_{\mathrm{conflict}}}$,
where the final token positions may differ because the conflict passage changes the prompt length. At each layer $i$, we compute

\begin{equation}
s_{T_{\mathrm{clean}},T_{\mathrm{conflict}}}^{\,i} =
\frac{\bar{h}^{(\mathrm{clean})\,i}_{T_{\mathrm{clean}}} \cdot \bar{h}^{(\mathrm{conflict})\,i}_{T_{\mathrm{conflict}}}}
     {\|\bar{h}^{(\mathrm{clean})\,i}_{T_{\mathrm{clean}}}\| \; \|\bar{h}^{(\mathrm{conflict})\,i}_{T_{\mathrm{conflict}}}\|}.
\label{eq:cosine}
\end{equation}

where $s_{T_{\mathrm{clean}},T_{\mathrm{conflict}}}^{\,i}$ is the cosine similarity between the two final positions. We select the layer with the lowest similarity:

\begin{equation}
\ell = \arg\min_i s_{T_{\mathrm{clean}},T_{\mathrm{conflict}}}^{\,i}.
\label{eq:layer_selection}
\end{equation}

This criterion requires no sweep over downstream task scores. The final layer is also excluded when its separation reflects the already-decoded output distribution rather than a reusable source-selection representation; only an earlier angular minimum is eligible for intervention.

The layer-wise curves in Appendix~\ref{sec:app-abl-layer} verify that the selected layer is the point of greatest clean--conflict separation across the evaluated models.

\paragraph{Extracting the direction.} At the selected layer $\ell$, we define the steering direction as a
difference-in-means contrast between the conflict and clean activations at the last token:

\begin{equation}
v_{T_{\mathrm{clean}},T_{\mathrm{conflict}}}^{\ell}
=
\bar{h}^{(\mathrm{clean})\,\ell}_{T_{\mathrm{clean}}}
-
\bar{h}^{(\mathrm{conflict})\,\ell}_{T_{\mathrm{conflict}}}.
\label{eq:direction}
\end{equation}

Thus $v_{\ell,\,T_{\mathrm{clean}},T_{\mathrm{conflict}}}$ points from the conflict regime toward stronger parameter-stored reliance.

\subsection{Applying the Steering Vector}
\label{sec:steer}

\paragraph{Inference-time intervention.}
We edit the \textit{resid post} stream only at the angular-separation-selected layer $\ell$ and the conflict prompt's final token. Appendix~\ref{sec:app-ablation} reports representation-hook, layer, and token-position ablations. Let $h_{T_{\mathrm{conflict}}}^{\ell}$ denote this activation. We apply the steering vector as

\begin{equation}
h'{}_{T_{\mathrm{conflict}}}^{\ell} \;=\; h_{T_{\mathrm{conflict}}}^{\ell} + \alpha\, v_{T_{\mathrm{clean}},T_{\mathrm{conflict}}}^{\ell}.
\label{eq:steer}
\end{equation}

Here, $\alpha$ is a signed control coefficient: positive values favor parametric reliance, negative values favor contextual reliance, and $\alpha=0$ recovers the unsteered conflict pass.

\subsection{Evaluation}
\label{sec:method-eval}

We measure steering by matching the first 256 generated tokens to the parametric and contextual answers. Each item receives one of four labels: \textit{parametric-only}, \textit{contextual-only}, \textit{both}, or \textit{*neither*}. Let $N$ be the number of evaluated items; the reported unconditional rates are
\[
\begin{aligned}
\mathrm{Acc}_{\mathrm{par}}&=\frac{N_{\mathrm{parametric\text{-}only}}}{N},\\
\mathrm{Acc}_{\mathrm{ctx}}&=\frac{N_{\mathrm{contextual\text{-}only}}}{N}.
\end{aligned}
\]
The main table retains these two directional rates. Appendix~\ref{sec:app-labeling} details normalization, the fixed answer window, the match rules, and the blinded-judge validation of the inexpensive string-match metric. We use train/validation/test splits for direction estimation, coefficient selection, and final reporting, with a 70/15/15 split for NQ-Swap.


\section{Continuous Control Along a Single Direction}

\label{sec:exp-onedim}
We first evaluate whether the model's knowledge selection behavior can be captured and continuously modulated along a single steering direction $v$. 
Here, the scalar $\alpha$ serves as a continuous control coefficient, with positive values shifting the model toward parametric reliance and negative values toward contextual reliance. 
Implementation details are provided in Appendix~\ref{app:implementation-details}.

\subsection{Improving Binary Source Choice}
\label{sec:binary-choice}
Table~\ref{tab:main} first shows that Ours improves the desired \textit{source-selection rate} over the unsteered model across the four models and both benchmarks, establishing control at a validation-selected operating point. The steering-strength sensitivity analysis in Appendix~\ref{app:main-alpha-stability} further shows stable transfer performance at neighboring $\alpha$ values around each selected operating point. Beyond these discrete endpoints, sweeping $\alpha$ reveals how the same direction changes source selection continuously.

\subsection{Continuous Control at Population and Instance Levels}
\label{sec:continuous-control}

Having established discrete source control at a validation-selected operating point, we next examine whether the same direction supports a continuous transition as $\alpha$ varies.


At the \textit{population level}, Figure~\ref{fig:main-curves} shows smooth, opposing changes in parametric and contextual selection as $\alpha$ varies across models and datasets. This aggregate trade-off is also visible within individual fixed examples, and Appendix~\ref{sec:large-models} shows that it extends to larger instruction-tuned models.


At the \textit{instance level}, Figure~\ref{fig:fig1} traces a response from one source through mixed outputs to the other as $\alpha$ changes. Thus, the aggregate trend reflects graded individual source selection rather than only a shift in proportions. Appendix~\ref{sec:app-four-model-cases} provides trajectories for every model and cases that do not reach the requested source.

\subsection{Robustness to External-Knowledge Strength and Question Difficulty}
\label{sec:robustness}
We test two factors: \textit{external-knowledge strength} and \textit{question difficulty}. As Table~\ref{tab:robustness-strength-main} shows, stronger contextual evidence raises unsteered contextual selection, while steering remains effective. Table~\ref{tab:robustness-difficulty-main} shows gains at every difficulty level, with the largest parametric gain on hard questions.

\begin{table}[htbp]
  \centering\footnotesize
  \caption{Robustness to external-knowledge strength on NQ-Swap with Llama-3.1-8B-Instruct (\%).}
  \label{tab:robustness-strength-main}
  \setlength{\tabcolsep}{2pt}
  \begin{tabularx}{\columnwidth}{@{}>{\centering\arraybackslash}X *{4}{>{\centering\arraybackslash}p{0.15\columnwidth}}@{}}
    \toprule
    {\scriptsize\shortstack{Context\\occurrences}} & {\scriptsize\shortstack{Unsteered\\parametric}} & {\scriptsize\shortstack{Parametric\\steering}} & {\scriptsize\shortstack{Unsteered\\contextual}} & {\scriptsize\shortstack{Contextual\\steering}} \\
    \midrule
    1 & 33.7 & 56.8 & 49.4 & 55.5 \\
    2 & 30.5 & 55.2 & 54.2 & 58.1 \\
    4 & 25.4 & 53.6 & 61.8 & 62.4 \\
    8 & 17.1 & 47.9 & 73.5 & 74.8 \\
    \bottomrule
  \end{tabularx}
\end{table}

\begin{table}[t]
  \centering\footnotesize
  \caption{Robustness to question difficulty on NQ-Swap with Llama-3.1-8B-Instruct (\%).}
  \label{tab:robustness-difficulty-main}
  \setlength{\tabcolsep}{2pt}
  \begin{tabularx}{\columnwidth}{@{}X *{4}{>{\centering\arraybackslash}p{0.15\columnwidth}}@{}}
    \toprule
    Difficulty & {\scriptsize\shortstack{Unsteered\\parametric}} & {\scriptsize\shortstack{Parametric\\steering}} & {\scriptsize\shortstack{Unsteered\\contextual}} & {\scriptsize\shortstack{Contextual\\steering}} \\
    \midrule
    Easy & 42.6 & 62.4 & 45.8 & 56.2 \\
    Medium & 33.1 & 56.1 & 49.6 & 57.3 \\
    Hard & 12.4 & 49.8 & 64.2 & 74.3 \\
    \bottomrule
  \end{tabularx}
\end{table}

\addtocounter{figure}{1}
\begin{figure*}[t]
  \centering
  \includegraphics[width=.64\textwidth]{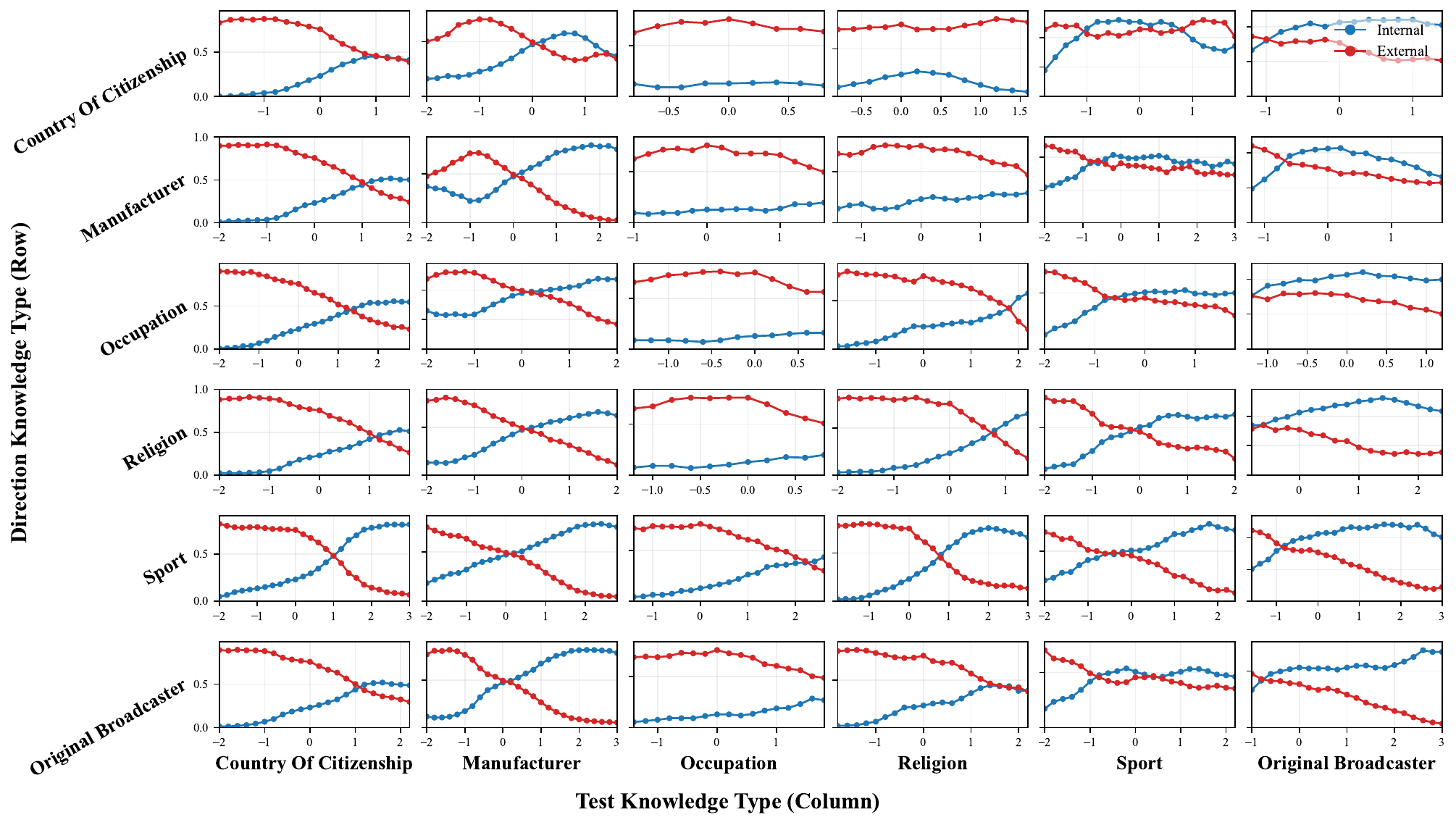}
  \caption{Controlled cross-relation transfer on CounterFact (Llama-3.1-8B-Instruct). Each cell evaluates a direction learned from one factual relation on another relation.}
  \label{fig:main-cross-relation}
\end{figure*}
\subsection{Knowledge Selection Is More Than Context Sensitivity}
\label{sec:selection-vs-context-attention}
The preceding results establish source control, but not whether it reflects explicit knowledge conflict or generic context sensitivity. We therefore compare source competition with matched prompts that add answer-aligned or irrelevant context without the same parametric--contextual conflict.

Table~\ref{tab:context-attention-control} compares no-context contrasts with conflicting, aligned, and irrelevant contexts. The conflicting-context contrast yields the strongest source-selection effects across datasets, whereas matched controls are weaker, indicating more than generic context sensitivity; Appendix~\ref{sec:app-context-cosine} further shows that the directions are similar but not identical (cosine 0.875--0.899).

\begin{table}[t]
  \centering
  \footnotesize
  \caption{Matched controls for knowledge selection versus generic context sensitivity (\%). The three steered columns contrast no context with conflicting, aligned, and irrelevant context.}
  \label{tab:context-attention-control}
  \setlength{\tabcolsep}{1.5pt}
  \renewcommand{\arraystretch}{1.03}
  \begin{tabular}{@{}lcccc@{}}
    \toprule
    Dataset & Unsteered & \shortstack{Conflicting\\context} & \shortstack{Aligned\\context} & \shortstack{Irrelevant\\context} \\
    \midrule
    \multicolumn{5}{@{}l@{}}{\textit{Panel A: Parametric-answer selection (positive $\alpha$)}} \\
    ParaConflict & 15.9 & \textbf{86.5} & 72.6 & 74.0 \\
    MacNoise     & 46.2 & \textbf{73.5} & 67.1 & 62.1 \\
    MemoTrap     & 67.1 & \textbf{84.1} & 71.7 & 75.3 \\
    CounterFact  & 36.0 & \textbf{54.4} & 46.1 & 49.1 \\
    \midrule
    \multicolumn{5}{@{}l@{}}{\textit{Panel B: Context/distractor-answer selection (negative $\alpha$)}} \\
    ParaConflict & 81.7 & \textbf{96.5} & 84.4 & 83.2 \\
    MacNoise     & 27.4 & \textbf{44.5} & 31.2 & 34.0 \\
    MemoTrap     & 29.7 & \textbf{57.0} & 49.9 & 47.5 \\
    CounterFact  & 60.8 & \textbf{64.1} & 57.6 & 55.1 \\
    \bottomrule
  \end{tabular}
\end{table}

\subsection{Preserving Performance Without Conflict}
\label{sec:nonconflict-preservation}
Finally, we test the cost of this source preference when no conflict is present. Table~\ref{tab:nonconflict-preservation-main} shows that, with the same fixed directions and coefficients, both directions retain substantial original-answer accuracy across all four datasets, although contextual steering incurs a larger trade-off on some datasets.

\begin{table}[t]
  \centering
  \footnotesize
  \caption{Preservation of the original answer without a conflicting context. Values are target-answer accuracy (\%); the same fixed parametric and contextual steering directions used in transfer are applied.}
  \label{tab:nonconflict-preservation-main}
  \setlength{\tabcolsep}{2.5pt}
  \renewcommand{\arraystretch}{1.05}
  \begin{tabular}{@{}lccc@{}}
    \toprule
    Dataset & Unsteered & \shortstack{Parametric\\direction} & \shortstack{Contextual\\direction} \\
    \midrule
    ParaConflict & 100.0 & 98.0 & 85.5 \\
    MacNoise & 100.0 & 80.8 & 86.7 \\
    MemoTrap & 100.0 & 84.2 & 91.1 \\
    CounterFact & 100.0 & 88.8 & 91.7 \\
    \bottomrule
  \end{tabular}
\end{table}

\section{Generalization of the Knowledge-Selection Direction}
\label{sec:generalization}

We next test whether this fixed direction survives distribution shifts. Section~\ref{sec:generalization-simple} tests simple factual-conflict tasks; Section~\ref{sec:generalization-complex} turns to complex tasks with external knowledge to acquire.

\setcounter{figure}{4}
\begin{figure}[tbp]
  \centering
  \includegraphics[width=.90\columnwidth]{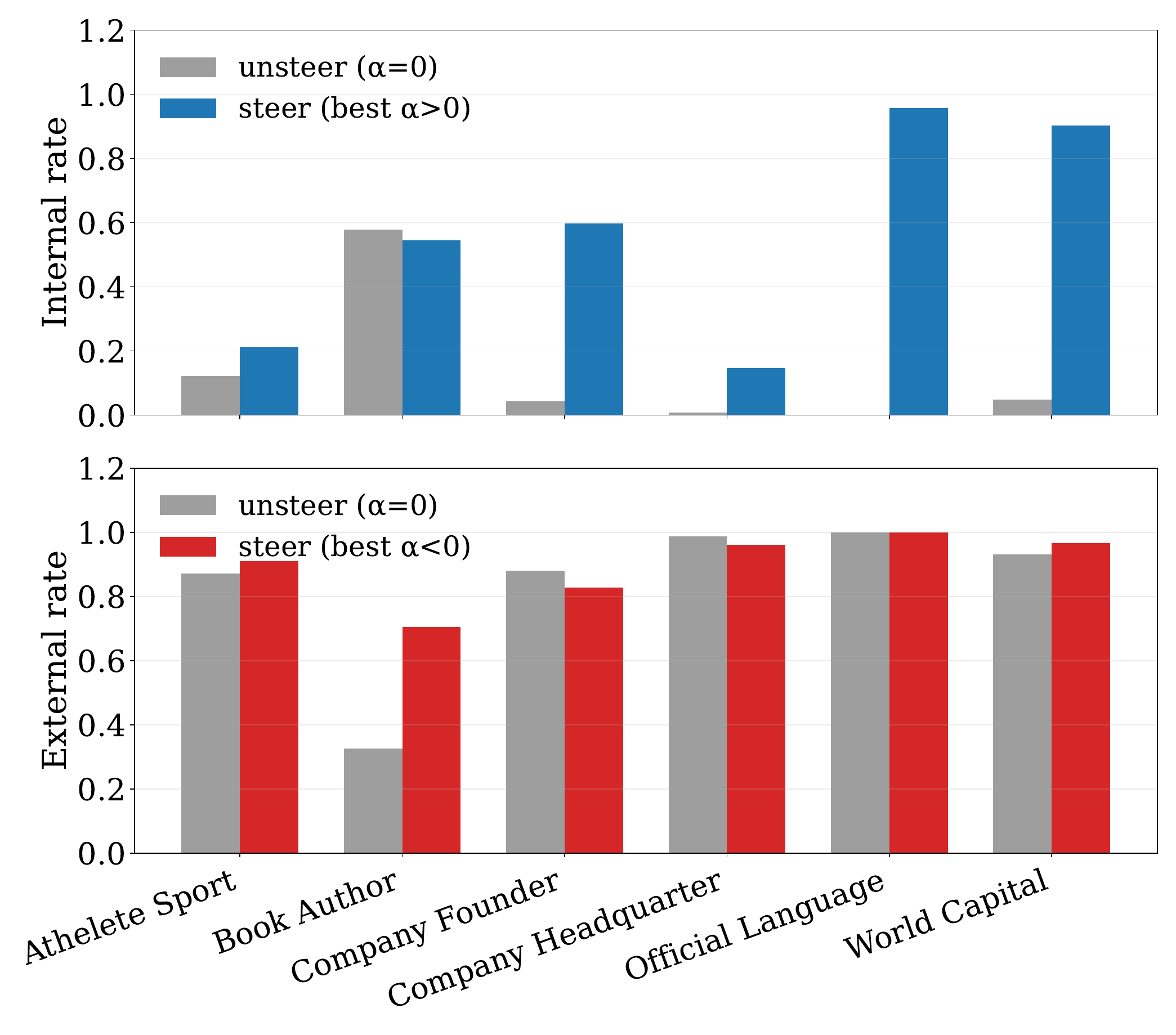}
  \caption{Controlled cross-format transfer on ParaConflict (Llama-3.1-8B-Instruct). The direction is fit on substitution-format conflicts and evaluated on coherent-format test items.}
  \label{fig:main-cross-format}
\end{figure}

\subsection{Across Controlled Simple Factual-Conflict Tasks}
\label{sec:generalization-simple}
These tests vary \textit{format}, \textit{semantic relation}, or \textit{benchmark construction} while retaining factual conflicts: prompts with competing answers. Thus, this level establishes generalization within the task family while retaining bidirectional steering.
\newcommand{\MainGeneralizationTable}{%
\begin{table*}[!t]
  \centering
  \scriptsize
  \caption{Generalization to complex tasks through external-knowledge steering (\%). Models are grouped vertically and tasks horizontally. ContextFocus, JUICE, and CK-PLUG are included as additional Llama-3.1-8B-Instruct baselines.}
  \label{tab:generalization-summary}
  \setlength{\tabcolsep}{4pt}
  \renewcommand{\arraystretch}{1.05}
  \begin{tabular}{@{}ll*{7}{c}@{}}
    \toprule
    \textbf{Model} & \textbf{Method} & \multicolumn{2}{c}{\cellcolor{ragUnsteer}\textbf{RAG}} & \multicolumn{3}{c}{\cellcolor{parUnsteer}\textbf{Memory}} & \multicolumn{2}{c}{\cellcolor{ctxUnsteer}\textbf{Rule}} \\
    \cmidrule(lr){3-4}\cmidrule(lr){5-7}\cmidrule(lr){8-9}
      & & RGB & CRAG & LoCoMo & LoCoMo-Plus & LongMemEval & IFEval & FollowBench \\
    \midrule
    \rowcolor{tablerow}\cellcolor{white}Llama-3.1-8B-Instruct & Unsteered & 62.0$\pm$1.2 & 23.7$\pm$0.8 & 8.5$\pm$0.4 & 32.7$\pm$1.1 & 43.2$\pm$0.9 & 33.6$\pm$1.0 & 14.5$\pm$0.6 \\
    \rowcolor{parBase}\cellcolor{white} & Random split & 64.0$\pm$1.4 & 24.7$\pm$0.9 & 8.5$\pm$0.4 & 32.7$\pm$1.0 & 42.8$\pm$0.8 & 33.9$\pm$1.1 & 12.9$\pm$0.5 \\
    \rowcolor{parBase}\cellcolor{white} & ContextFocus & 69.0$\pm$0.3 & 24.5$\pm$1.8 & 10.6$\pm$1.8 & 34.9$\pm$2.1 & 44.1$\pm$0.9 & 33.4$\pm$1.7 & 13.1$\pm$1.3 \\
    \rowcolor{parBase}\cellcolor{white} & JUICE & 62.0$\pm$1.1 & 23.7$\pm$0.7 & 7.6$\pm$0.3 & 34.2$\pm$1.9 & 42.5$\pm$2.8 & 31.0$\pm$0.9 & 13.8$\pm$1.4 \\
    \rowcolor{parBase}\cellcolor{white} & CK-PLUG & 71.0$\pm$1.4 & 25.1$\pm$0.9 & 12.9$\pm$1.6 & 35.3$\pm$1.2 & 46.8$\pm$1.0 & 34.1$\pm$1.1 & 14.6$\pm$0.6 \\
    \rowcolor{parOurs}\cellcolor{white} & Ours & \textbf{82.0}$\pm$0.5 & \textbf{25.8}$\pm$1.0 & \textbf{13.2}$\pm$0.9 & \textbf{36.7}$\pm$1.3 & \textbf{45.4}$\pm$1.1 & \textbf{35.0}$\pm$1.2 & \textbf{14.9}$\pm$0.7 \\
    \midrule
    \rowcolor{tablerow}\cellcolor{white}Qwen3-8B & Unsteered & 72.0$\pm$1.4 & 27.5$\pm$1.0 & 9.8$\pm$0.5 & 28.2$\pm$0.9 & 42.8$\pm$0.9 & 34.9$\pm$1.1 & 15.7$\pm$0.5 \\
    \rowcolor{parBase}\cellcolor{white} & Random split & 72.0$\pm$1.1 & 27.5$\pm$0.5 & 10.1$\pm$0.5 & 28.5$\pm$1.0 & 42.1$\pm$0.4 & 34.9$\pm$1.0 & 15.7$\pm$1.3 \\
    \rowcolor{parBase}\cellcolor{white} & ContextFocus & 72.1$\pm$0.8 & 27.6$\pm$1.3 & 9.9$\pm$0.6 & 28.9$\pm$1.1 & 42.9$\pm$0.4 & 34.8$\pm$1.1 & 16.6$\pm$1.9 \\
    \rowcolor{parBase}\cellcolor{white} & JUICE & 68.5$\pm$1.2 & 25.4$\pm$1.8 & 9.0$\pm$0.4 & 26.8$\pm$0.8 & 41.6$\pm$0.8 & 33.1$\pm$0.9 & 14.2$\pm$1.2 \\
    \rowcolor{parBase}\cellcolor{white} & CK-PLUG & 74.3$\pm$0.5 & 28.3$\pm$1.1 & 10.6$\pm$0.6 & 33.5$\pm$1.2 & 44.1$\pm$1.0 & 35.4$\pm$1.2 & 16.6$\pm$0.8 \\
    \rowcolor{parOurs}\cellcolor{white} & Ours & \textbf{79.0}$\pm$1.6 & \textbf{30.0}$\pm$1.2 & \textbf{13.1}$\pm$0.7 & \textbf{45.4}$\pm$1.5 & \textbf{43.8}$\pm$1.1 & \textbf{36.2}$\pm$1.3 & \textbf{18.3}$\pm$0.9 \\
    \midrule
    \rowcolor{tablerow}\cellcolor{white}Gemma-2-9B-IT & Unsteered & 71.0$\pm$1.3 & 26.6$\pm$0.9 & 8.8$\pm$0.4 & 9.2$\pm$0.3 & 42.0$\pm$0.7 & 29.2$\pm$0.9 & 16.1$\pm$0.1 \\
    \rowcolor{parBase}\cellcolor{white} & Random split & 72.0$\pm$0.7 & 26.9$\pm$0.9 & 8.8$\pm$0.5 & 11.0$\pm$0.4 & 41.7$\pm$0.9 & 29.8$\pm$1.2 & 15.7$\pm$0.7 \\
    \rowcolor{parBase}\cellcolor{white} & ContextFocus & 77.0$\pm$1.1 & 26.9$\pm$1.2 & 8.7$\pm$0.2 & 11.1$\pm$1.2 & 42.1$\pm$1.3 & 29.7$\pm$0.7 & 16.1$\pm$1.8 \\
    \rowcolor{parBase}\cellcolor{white} & JUICE & 75.0$\pm$1.2 & 25.0$\pm$0.6 & 8.2$\pm$0.7 & 9.6$\pm$0.3 & 41.1$\pm$0.7 & 27.9$\pm$0.3 & 15.0$\pm$0.6 \\
    \rowcolor{parBase}\cellcolor{white} & CK-PLUG & 81.0$\pm$0.9 & 27.0$\pm$1.0 & 9.3$\pm$0.5 & 14.4$\pm$0.9 & 43.6$\pm$1.4 & 30.0$\pm$1.1 & 16.5$\pm$2.3 \\
    \rowcolor{parOurs}\cellcolor{white} & Ours & \textbf{88.0}$\pm$1.1 & \textbf{27.7}$\pm$1.1 & \textbf{10.5}$\pm$0.6 & \textbf{17.2}$\pm$0.6 & \textbf{44.2}$\pm$1.2 & \textbf{30.5}$\pm$0.9 & \textbf{17.0}$\pm$0.8 \\
    \midrule
    \rowcolor{tablerow}\cellcolor{white}Mistral-7B-Instruct-v0.3 & Unsteered & 81.0$\pm$1.5 & 27.3$\pm$1.0 & 9.4$\pm$0.5 & 32.7$\pm$1.1 & 40.2$\pm$0.8 & 25.1$\pm$0.8 & 14.9$\pm$0.4 \\
    \rowcolor{parBase}\cellcolor{white} & Random split & 81.0$\pm$1.4 & 27.1$\pm$0.9 & 9.1$\pm$0.4 & 33.2$\pm$1.2 & 39.4$\pm$0.9 & 25.1$\pm$0.9 & 14.9$\pm$0.6 \\
    \rowcolor{parBase}\cellcolor{white} & ContextFocus & 82.0$\pm$0.5 & 27.8$\pm$1.0 & 9.5$\pm$0.7 & 34.6$\pm$1.7 & 40.1$\pm$0.8 & 23.3$\pm$1.3 & 14.8$\pm$1.3 \\
    \rowcolor{parBase}\cellcolor{white} & JUICE & 77.0$\pm$1.7 & 25.8$\pm$0.8 & 8.7$\pm$1.4 & 31.9$\pm$1.0 & 39.3$\pm$0.7 & 23.6$\pm$0.7 & 13.7$\pm$0.5 \\
    \rowcolor{parBase}\cellcolor{white} & CK-PLUG & 82.0$\pm$0.6 & 28.2$\pm$1.1 & 9.9$\pm$0.4 & 33.1$\pm$1.3 & 41.6$\pm$1.5 & 26.4$\pm$1.0 & 15.3$\pm$1.1 \\
    \rowcolor{parOurs}\cellcolor{white} & Ours & \textbf{86.0}$\pm$1.2 & \textbf{28.5}$\pm$0.6 & \textbf{10.7}$\pm$0.6 & \textbf{35.9}$\pm$1.3 & \textbf{41.6}$\pm$1.0 & \textbf{29.4}$\pm$1.1 & \textbf{15.9}$\pm$0.8 \\
    \bottomrule
  \end{tabular}
\end{table*}
}

\begin{table}[!b]
  \centering
  \footnotesize
  \caption{Controlled factual-conflict cross-dataset transfer on Llama-3.1-8B-Instruct (\%). The direction and coefficients are selected only on NQ-Swap.}
  \label{tab:crossdata}
  \scriptsize
  \setlength{\tabcolsep}{3pt}
  \renewcommand{\arraystretch}{0.88}
  \begin{tabular}{@{}lcc|cc@{}}
    \toprule
    & \multicolumn{2}{c}{Parametric steering} & \multicolumn{2}{c}{Contextual steering} \\
    \cmidrule(lr){2-3}\cmidrule(lr){4-5}
    Test data & Unsteered & Ours & Unsteered & Ours \\
    \midrule
    MacNoise & 46.2 & \textbf{60.0} & 27.0 & \textbf{34.1} \\
    ParaConflict & 15.8 & \textbf{83.9} & 81.4 & \textbf{90.8} \\
    CounterFact & 35.7 & \textbf{38.9} & 60.0 & \textbf{60.7} \\
    MemoTrap & 66.7 & \textbf{75.4} & 30.0 & \textbf{53.3} \\
    \bottomrule
\end{tabular}
\end{table}

\MainGeneralizationTable

\noindent\textbf{Cross-format transfer.} We begin by varying the \textit{conflict format} while holding the fact and distractor fixed. ParaConflict separates a short substitution conflict from a richer coherent conflict for the same fact; Appendix~\ref{app:prompt-templates} gives the full prompt forms. Both prompts inject the same distractor, ``football,'' for the correct answer ``basketball'':

\begin{tcolorbox}[colback=gray!10, colframe=gray!80, left=3pt, right=3pt, top=2pt, bottom=2pt]
\small\textbf{Short:} \textit{Michael Jordan plays the sport of football.}
\end{tcolorbox}
\begin{tcolorbox}[colback=gray!10, colframe=gray!80, left=3pt, right=3pt, top=2pt, bottom=2pt]
\small\textbf{Coherent:} \textit{Recognized by peers and fans alike, Michael Jordan's journey in football has been highlighted in sports publications and athlete profiles. $\ldots$ The impact of Michael Jordan in football is celebrated in community events and athletic programs. $\ldots$}
\end{tcolorbox}

Figure~\ref{fig:main-cross-format} shows that a direction fitted on the short format also changes source selection in the coherent format, so the effect is not tied to the estimation template. Appendix~\ref{app:cross-format} reports the corresponding results for the other models.

\noindent\textbf{Cross-relation transfer.} Having shown that the direction is not tied to one prompt template, we next hold the conflict structure fixed and vary the \textit{semantic relation}. CounterFact lets us learn directions from \textit{country of citizenship}, \textit{manufacturer}, \textit{occupation}, \textit{religion}, \textit{sport}, and \textit{original broadcaster}, then evaluate each direction on the other relations. This tests whether the direction encodes source selection rather than lexical or semantic features of one relation. The off-diagonal effects in Figure~\ref{fig:main-cross-relation} show transfer across all six knowledge categories; Appendix~\ref{app:cross-knowledge-type} provides the other-model heatmaps.

\noindent\textbf{Cross-dataset transfer.} We apply the direction learned on NQ-Swap to distinct factual-conflict datasets without target-side refitting. These dataset shifts introduce variations in conflict construction, factual content, and other nuisance factors, reducing the possibility that the observed effect is tied to dataset-specific properties. Table~\ref{tab:crossdata} demonstrates bidirectional transfer across datasets, while context-only controls achieve consistently weaker effects.

\subsection{To Complex, Application-Relevant Tasks}
\label{sec:generalization-complex}

We next evaluate whether a direction learned from explicit factual conflicts transfers to broader input-grounded tasks without target-task direction fitting. Unlike the controlled benchmarks, these tasks do not necessarily involve direct competition between parametric and contextual answers.

\paragraph{RAG.} RAG tasks provide \textit{retrieved documents}: RGB supplies deliberately counterfactual evidence, whereas CRAG uses retrieved passages that can be long, stale, noisy, or only partly relevant. Success requires selecting the supplied evidence when it conflicts with a plausible parametric answer.

\paragraph{Memory.} Memory tasks provide \textit{multi-turn histories}. In LoCoMo, LoCoMo-Plus, and LongMemEval, the relevant fact may lie many turns from the question and compete with a continuation, stale association, or another event. These tasks test retrieval, localization, and temporal ordering of knowledge rather than passage matching.

\paragraph{Rule-following.} Rule-following tasks provide \textit{explicit behavioral constraints}. IFEval and FollowBench require the model to retain rules about output structure, length, required content, and exclusions throughout generation; they serve as instruction-following controls for whether the direction improves use of contextual constraints.

\paragraph{Results.} Table~\ref{tab:generalization-summary} reports held-out unsteered and steered scores using source matching for controlled factual tasks and each benchmark's native evaluator otherwise. The corresponding item-level changes and error-stratified cases are reported in Appendix~\ref{sec:app-steering-audit}, including the case tables.

\section{Conclusion}

We show that a single residual-stream direction enables smooth, bidirectional selection between parametric knowledge and external context. Changing its signed coefficient at one carefully selected layer shifts both aggregate source-match rates and fixed-instance responses.

The control generalizes across conflict formats, factual relations, and datasets, and transfers to RAG, multi-turn memory, and rule-following tasks without target-task direction fitting. Weaker context-attention controls suggest that transfer reflects source competition. These results provide evidence for a transferable control direction underlying knowledge-source selection.\section*{Limitations}
\label{sec:limitations}

First, our method requires white-box access to model activations.
Since the direction is estimated and applied in the residual stream, it is not directly applicable to closed-source or API-only models.

Second, the learned direction should be viewed as a useful control axis rather than a complete mechanism of knowledge selection.
Although a single direction enables effective steering, source selection may still involve broader circuits, including attention heads, MLP features, and decoding dynamics.

Third, our evaluation is based mainly on controlled factual-conflict benchmarks and source-choice accuracy.
This setup isolates parametric versus contextual reliance, but may not capture more nuanced outputs such as hedging, partial agreement, or answers that combine both sources.

Finally, the direction improves the complex tasks we evaluate, but the gains are often smaller than in controlled factual conflicts and depend on the underlying capabilities of each model. Establishing how model competence, task difficulty, and evidence quality jointly determine these effects remains future work.
\section*{Acknowledgments}
This work is supported by CCF-Tencent Rhino-Bird Open Research Fund (Grant No. CCF-Tencent RAGR20260112).

\bibliography{custom}


\newpage
\listofappendices

\begin{appendices}

\section{Dataset Details}
\label{app:dataset-details}

We briefly describe how each benchmark is converted into our source-choice setting. 
For all datasets, we first keep examples where the model answers the clean prompt correctly. 
We then evaluate the same examples under a conflict prompt that supports an alternative answer. 
The original answer is treated as the \textit{parametric} answer, while the answer supported by the conflict prompt is treated as the \textit{contextual} answer. 
Prompt examples are provided in Appendix~\ref{app:prompt-templates}. 

\paragraph{NQ-Swap.}
{NQ-Swap} pairs Natural Questions examples with substituted passages. 
We use the original answer as the parametric answer and the substituted answer as the contextual answer. 
The clean prompt contains only the question, while the conflict prompt prepends the substituted context before the same question. 
We use the question-deduplicated split without tables or list items.

\paragraph{MacNoise.}
{MacNoise} provides QA examples with perturbed retrieved contexts. 
The original answer is used as the parametric answer, and the replacement answer is used as the contextual answer. 
For each example, we select the first perturbed context marked by \texttt{pert\_lbls\_100 == 1} and prepend its \texttt{text\_pert} field to form the conflict prompt.

\paragraph{ParaConflict.}
{ParaConflict} contains synthetic factual conflicts across multiple categories. 
The gold answer is treated as the parametric answer, and the distracted token is treated as the contextual answer. 
We use the coherent conflict format in the main experiments because it provides a natural-language passage supporting the distractor. 
For cross-format transfer, we train on substitution-format items and evaluate on coherent-format items.

\paragraph{CounterFact.}
{CounterFact} contains factual editing examples grouped by Wikidata relation. 
The true target is treated as the parametric answer, and the edited target is treated as the contextual answer. 
The conflict prompt is constructed by inserting the edited target into a paraphrased prefix and appending the original factual query. 
For cross-knowledge-type transfer, we evaluate six relations: country of citizenship, manufacturer, occupation, religion, sport, and original broadcaster.

\paragraph{MemoTrap.}
{MemoTrap} creates conflicts between memorized phrase completions and explicit prompt instructions. 
The familiar phrase continuation is treated as the parametric answer, while the instruction-satisfying completion is treated as the contextual answer. 
The clean prompt removes the explicit ending constraint, whereas the conflict prompt keeps the original instruction.

\paragraph{RGB and CRAG.}
RGB supplies retrieved, deliberately counterfactual passages whose answer conflicts with a familiar factual answer; we evaluate whether the output follows the supplied evidence. CRAG supplies retrieved web-style passages for factual questions, including cases with stale, competing, noisy, or partially relevant evidence. We use each benchmark's held-out evaluation items and its factual source-match protocol.

\paragraph{LoCoMo, LoCoMo-Plus, and LongMemEval.}
These datasets require answers grounded in multi-turn personal histories. LoCoMo and LoCoMo-Plus test retrieval and linkage across earlier dialogue turns, while LongMemEval additionally stresses temporal ordering and aggregation across events. We preserve each benchmark's native held-out split and exact-match evaluator.

\paragraph{IFEval and FollowBench.}
These instruction-following datasets specify explicit requirements on output structure, length, content, or lexical exclusions. We use their native constraint-compliance evaluators; they serve as controls for retaining and applying external instructions across the generated response.

\section{Intervention Location and Representation Validation}
\label{sec:app-ablation}

We ablate each modeling choice underlying our main pipeline ({Llama-3.1-8B-Instruct}, {NQ-Swap} test unless noted). For each variant we re-estimate $\hat{v}$ and $\ell$ when required, re-tune $(\alpha^{+},\alpha^{-})$ on validation, and report Acc$_\mathrm{par}$ and Acc$_\mathrm{ctx}$ on the test split.

\subsection{Representation Hook}
\label{sec:app-abl-representation}

\begin{table}[tbp]
    \centering
    \footnotesize
    \setlength{\tabcolsep}{8pt}
    \renewcommand{\arraystretch}{1.08}
    \caption{Steered source-choice accuracy (\%) on {NQ-Swap} test across all evaluated models.
    Acc$_\mathrm{par}$ and Acc$_\mathrm{ctx}$ report source-choice accuracy when steering toward parametric and contextual knowledge, respectively.
    Each steered cell reports \textit{resid-mid} / \textit{resid-post} (best $\alpha$ per direction).}
    \label{tab:app-steered}
    \begin{tabular}{@{}l c c@{}}
        \toprule
        Model & Acc$_\mathrm{par}$ & Acc$_\mathrm{ctx}$ \\
        \midrule
        Llama-3.1-8B-Instruct  & 50.4\,/\,\textbf{56.8} & \textbf{59.7}\,/\,55.5 \\
        Qwen3-8B       & \textbf{62.4}\,/\,54.6 & 51.1\,/\,\textbf{56.0} \\
        Gemma-2-9B-IT  & 33.8\,/\,\textbf{80.9} & 53.2\,/\,\textbf{65.4} \\
        Mistral-7B-Instruct-v0.3 & \textbf{59.0}\,/\,59.0 & 71.1\,/\,\textbf{60.2} \\
        \bottomrule
    \end{tabular}
\end{table}
Table~\ref{tab:app-steered} shows that resid post yields stronger bidirectional control; resid mid generally underperforms, indicating that the MLP sublayer contributes to the separable knowledge-selection signal.
\subsection{Layer Selection}
\label{sec:app-abl-layer}

\begin{figure}[tbp]
  \centering
  \footnotesize
  \begin{minipage}[t]{0.48\linewidth}
    \centering
    \includegraphics[width=\linewidth]{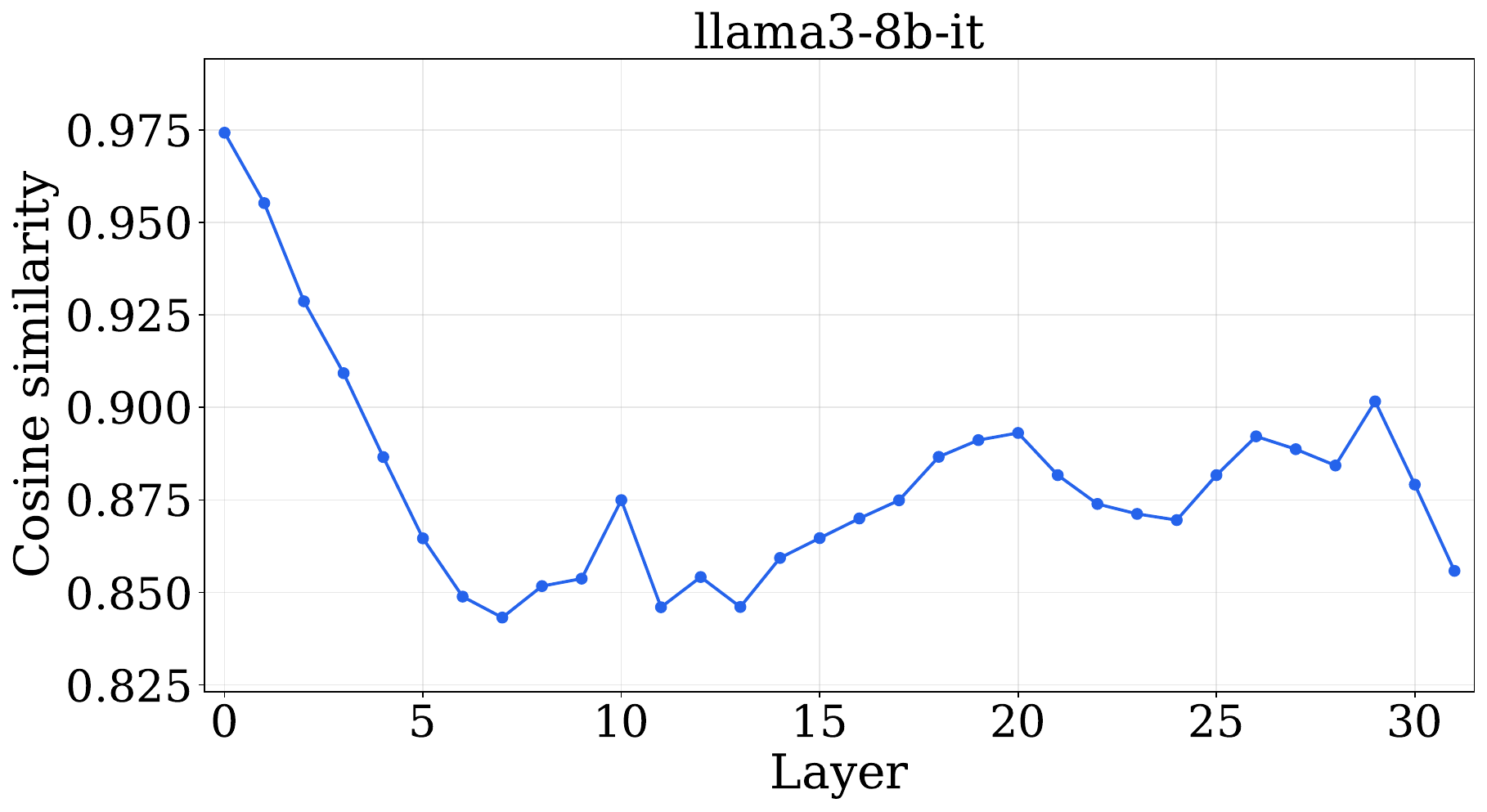}\\
    Llama-3.1-8B-Instruct
  \end{minipage}\hfill
  \begin{minipage}[t]{0.48\linewidth}
    \centering
    \includegraphics[width=\linewidth]{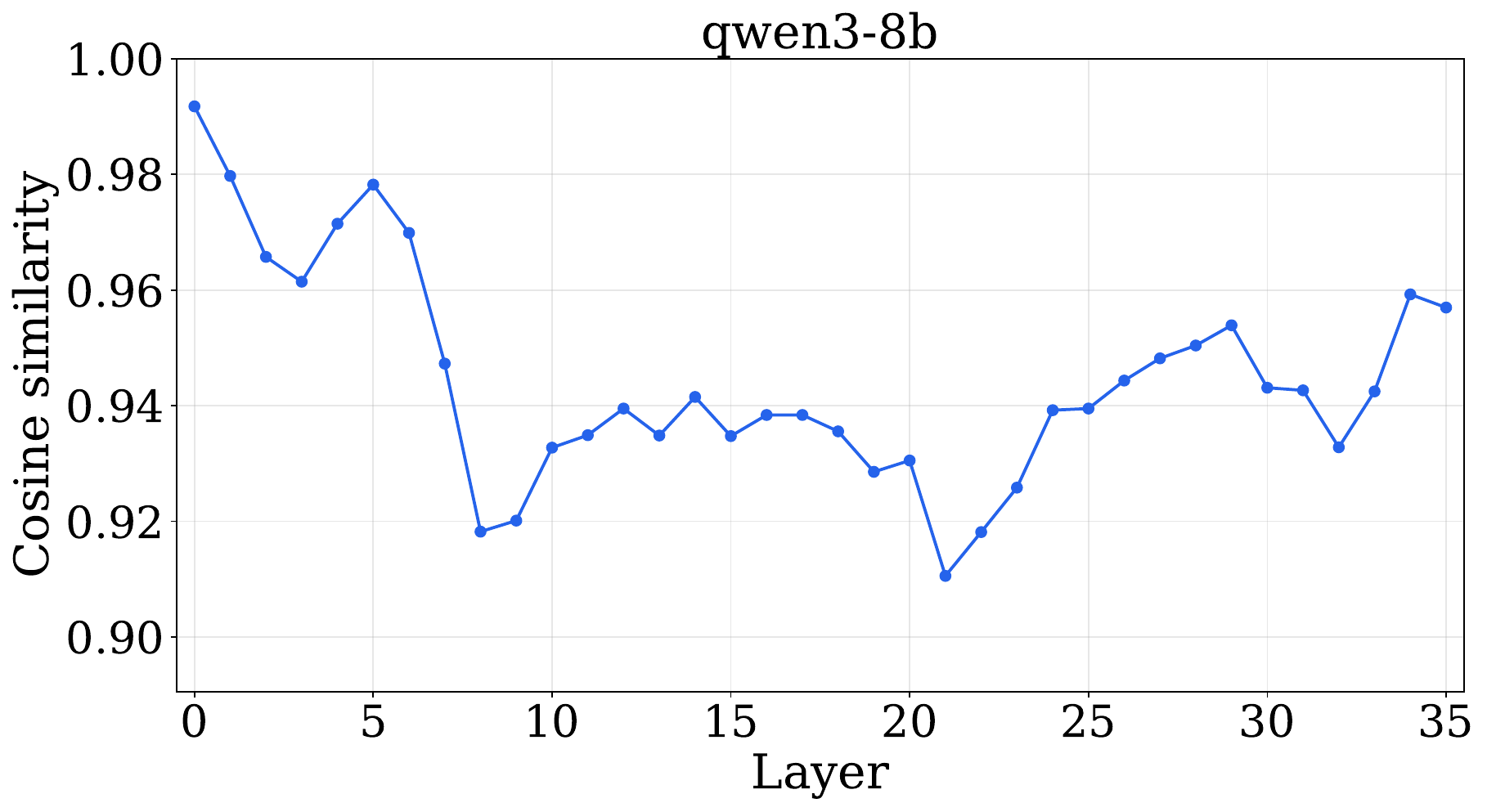}\\
    Qwen3-8B
  \end{minipage}\\[0.5em]
  \begin{minipage}[t]{0.48\linewidth}
    \centering
    \includegraphics[width=\linewidth]{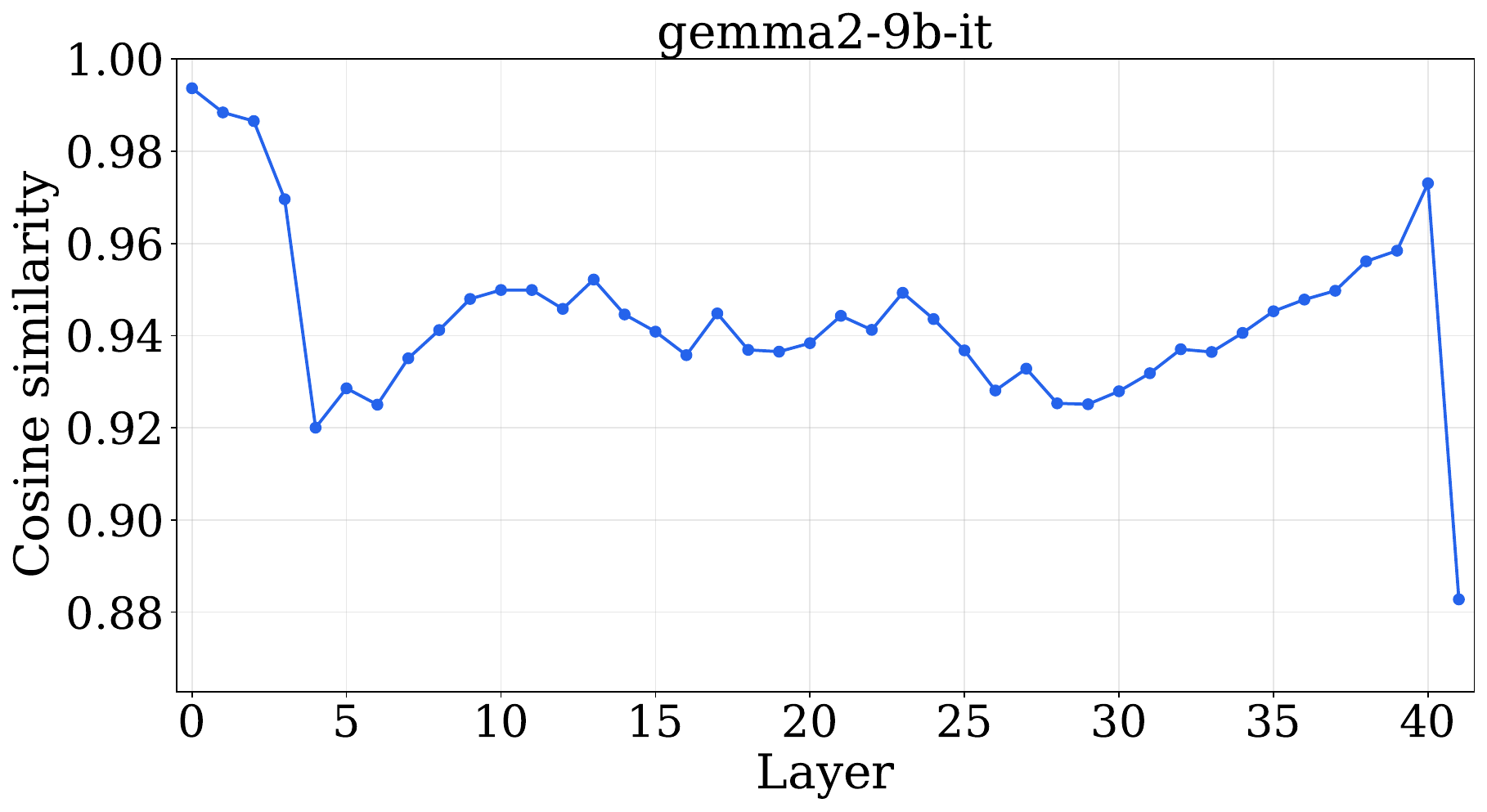}\\
    Gemma-2-9B-Instruct
  \end{minipage}\hfill
  \begin{minipage}[t]{0.48\linewidth}
    \centering
    \includegraphics[width=\linewidth]{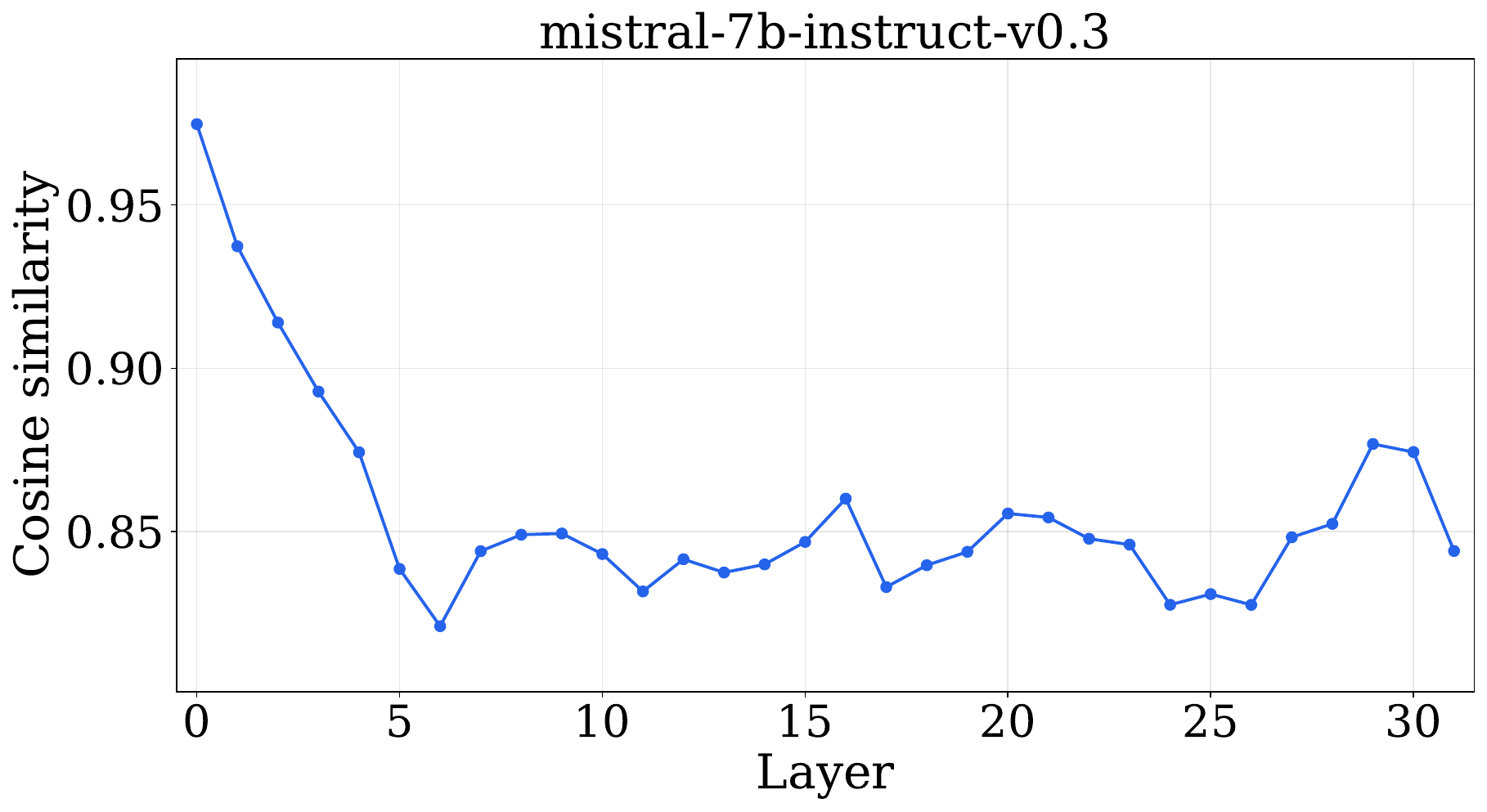}\\
    Mistral-7B-Instruct-v0.3
  \end{minipage}
  \caption{Layer-wise cosine similarity $s_{T_{\mathrm{clean}},T_{\mathrm{conflict}}}^{\,i}$ between the respective final-token activations of clean and conflict prompts (\eqref{eq:cosine}) on {NQ-Swap}.}
  \label{fig:angle-layers}
\end{figure}

\begin{figure*}[tbp]
  \centering
  \begin{subfigure}[t]{0.48\textwidth}
    \centering
    \includegraphics[width=\linewidth]{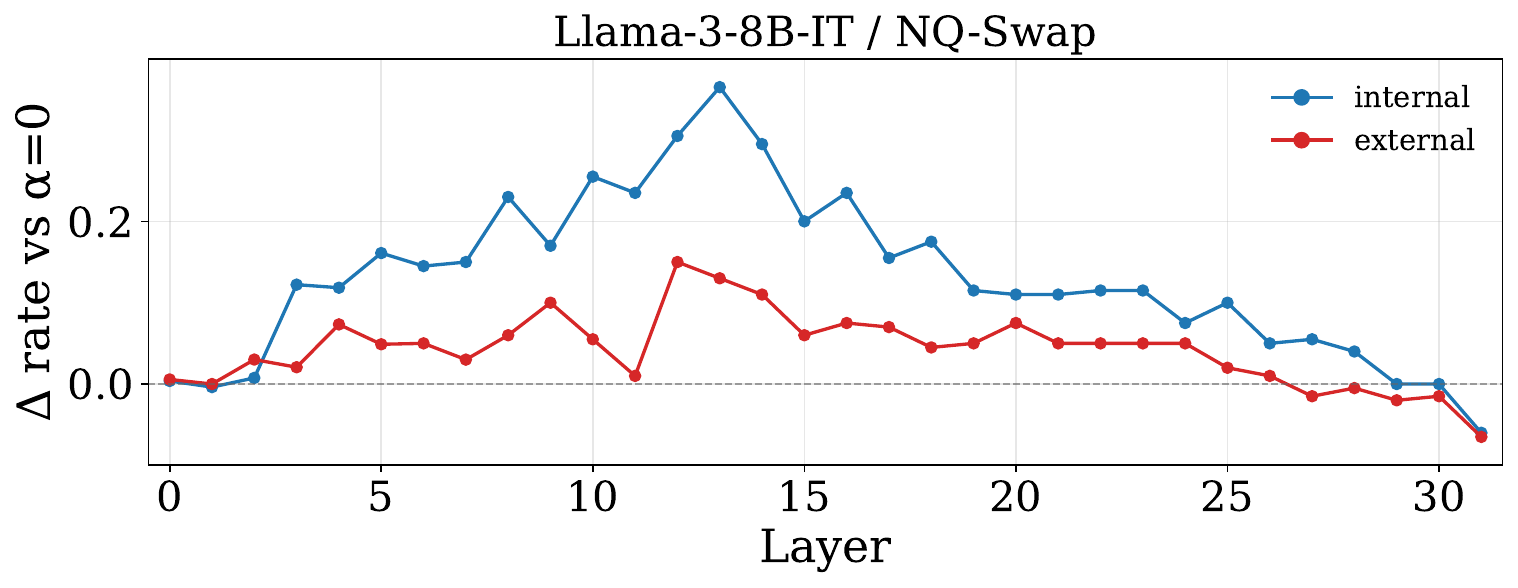}
    \caption{Llama-3.1-8B-Instruct}
  \end{subfigure}
  \hfill
  \begin{subfigure}[t]{0.48\textwidth}
    \centering
    \includegraphics[width=\linewidth]{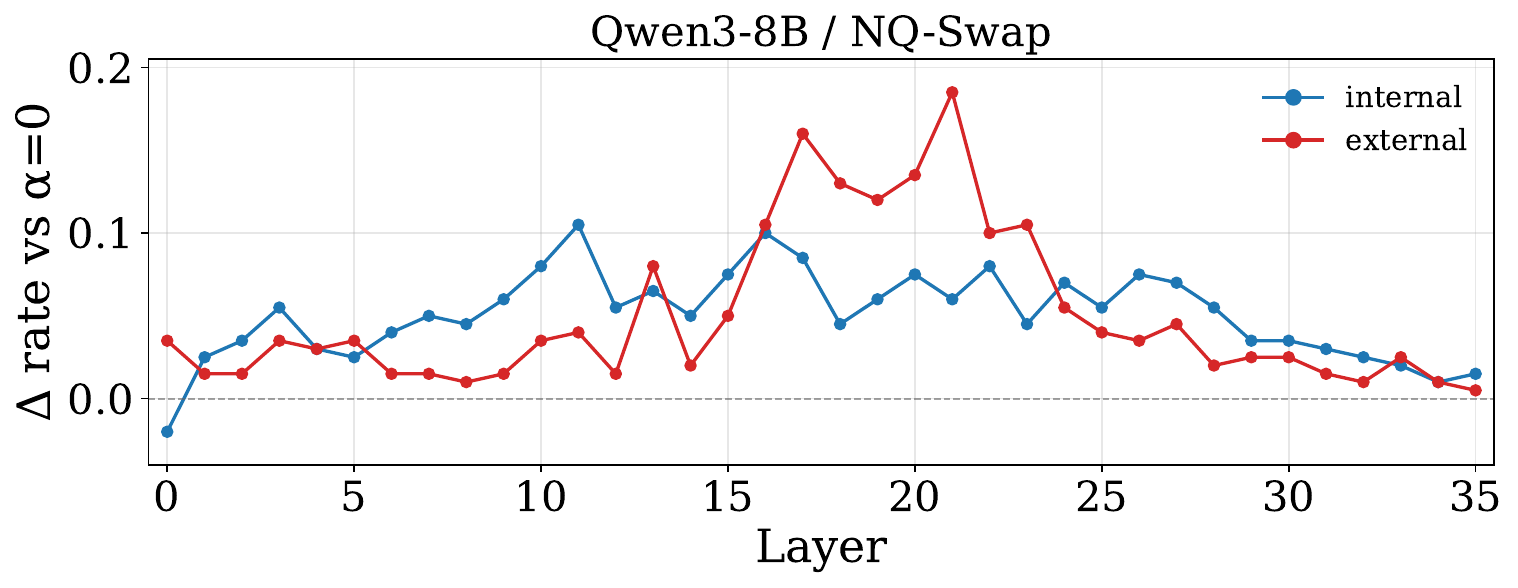}
    \caption{Qwen3-8B}
  \end{subfigure}

  \vspace{0.6em}

  \begin{subfigure}[t]{0.48\textwidth}
    \centering
    \includegraphics[width=\linewidth]{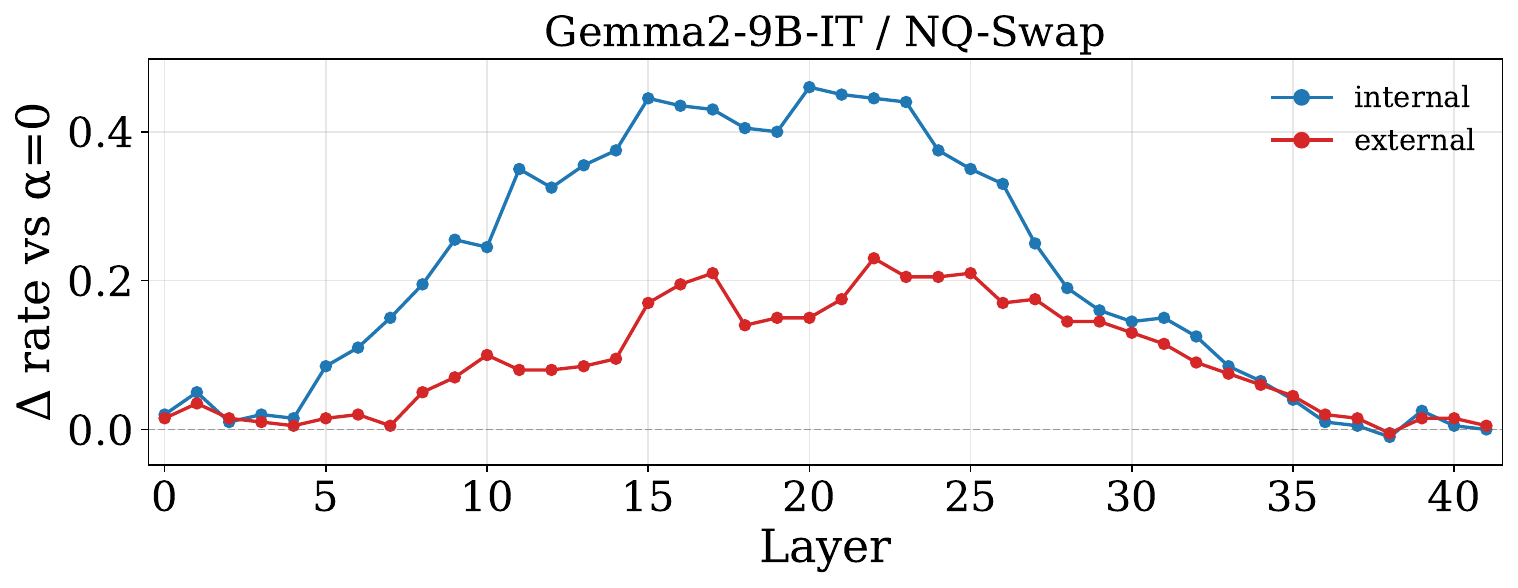}
    \caption{Gemma-2-9B-IT}
  \end{subfigure}
  \hfill
  \begin{subfigure}[t]{0.48\textwidth}
    \centering
    \includegraphics[width=\linewidth]{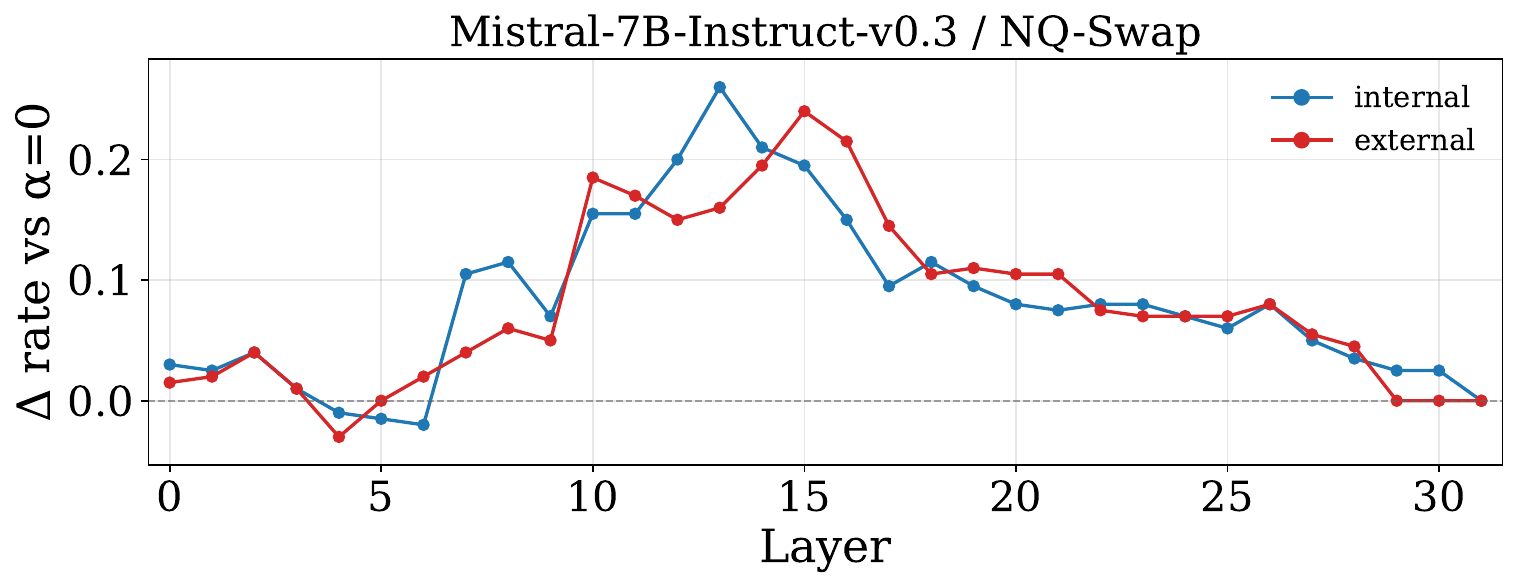}
    \caption{Mistral-7B-Instruct-v0.3}
  \end{subfigure}
  \caption{Layer-selection analysis on {NQ-Swap}. Each panel shows performance when the intervention is applied at different layers for one model. The layer-wise performance trend is consistent with the cosine-based separation pattern: layers with lower clean--conflict cosine similarity, corresponding to larger angular separation, tend to produce stronger steering effects.}
  \label{fig:app-abl-layer}
\end{figure*}

Instead of steering only at the selected layer $\ell$, we inject the intervention at each layer $i$ separately and examine the resulting performance across all four models on {NQ-Swap}. 
Figure~\ref{fig:angle-layers} reports the cosine curves used for localization, while Figure~\ref{fig:app-abl-layer} shows Acc$_\mathrm{par}$ and Acc$_\mathrm{ctx}$ as a function of the intervention layer. 
Across models, the layer-wise steering performance follows a similar trend to the angular separation scores: layers with larger separation generally yield stronger steering effects, while shallow and very late layers show weaker or less stable control. 
This alignment supports our layer-selection strategy and suggests that the angular separation score is a useful indicator of where knowledge selection is most steerable.

Across the completed sweeps, peaks align with the angular-separation layer; shallow and late layers show weaker or asymmetric control.
\subsection{Context-Sensitivity Direction Similarity}
\label{sec:app-context-cosine}

Table~\ref{tab:context-contrast-cosine} reports cosine similarities among the three no-context contrasts used in the matched context-sensitivity controls. A denotes no context; B, C, and D denote aligned, conflicting, and irrelevant context.

\begin{table}[tbp]
  \centering
  \footnotesize
  \caption{Cosine similarity between the three matched no-context contrasts on NQ-Swap.}
  \label{tab:context-contrast-cosine}
  \setlength{\tabcolsep}{5pt}
  \renewcommand{\arraystretch}{1.08}
  \begin{tabular}{@{}lc@{}}
    \toprule
    Contrast pair & Cosine similarity \\
    \midrule
    A--C vs. A--B & 0.899399 \\
    A--C vs. A--D & 0.878706 \\
    A--B vs. A--D & 0.874608 \\
    \bottomrule
  \end{tabular}
\end{table}

\subsection{Token Position}
\label{sec:app-abl-token}

\begin{table}[tbp]
\centering
  \footnotesize
  \setlength{\tabcolsep}{8pt}
  \renewcommand{\arraystretch}{1.08}
  \caption{Token-position ablation on {NQ-Swap} test.
  Acc$_\mathrm{par}$ and Acc$_\mathrm{ctx}$ report source-choice accuracy when steering toward parametric and contextual knowledge, respectively.
  Each entry is \textit{question-final / answer-final} token, using validation-selected coefficients at that position.}
  \label{tab:app-abl-token}
  \begin{tabular}{@{}l c c@{}}
    \toprule
    Model & Acc$_\mathrm{par}$ & Acc$_\mathrm{ctx}$ \\
    \midrule
    Llama-3.1-8B-Instruct  & 56.8 / \textbf{65.6} & \textbf{55.5} / 54.0 \\
    Qwen3-8B       & \textbf{54.6} / 47.1 & \textbf{56.0} / 44.5 \\
    Gemma-2-9B-IT  & \textbf{80.9} / 75.3 & \textbf{65.4} / 62.3 \\
    Mistral-7B-Instruct-v0.3 & 59.0 / \textbf{63.9} & \textbf{60.2} / 57.8 \\
    \bottomrule
  \end{tabular}
\end{table}

As shown in Table~\ref{tab:app-abl-token}, the question-final token generally yields stronger bidirectional control than the answer-final token, although answer-final intervention can perform better for parametric steering in a few models.

\section{Judge-Based Evaluation Protocol}
\label{sec:app-labeling}

Acc$_\mathrm{par}$ and Acc$_\mathrm{ctx}$ depend on how free-form outputs are mapped to discrete source labels. This appendix makes the protocol fully explicit.

\subsection{Answer Normalization}
\label{sec:app-norm}

Each gold parametric answer $a_\mathrm{par}$ and contextual answer $a_\mathrm{ctx}$ is normalized before matching:
\begin{enumerate}[topsep=0pt,itemsep=1pt,parsep=0pt,leftmargin=*]
  \item lowercase;
  \item remove punctuation except intra-token hyphens;
  \item remove leading articles (\textit{a}, \textit{an}, \textit{the});
  \item collapse whitespace;
\end{enumerate}

\subsection{Answer Window}
\label{sec:app-answer-window}

We take the first \textbf{256 generated tokens} of the raw continuation and normalize this fixed answer window. This avoids making the metric depend on sentence segmentation or on whether steering changes the placement of a sentence boundary. Outputs shorter than 256 tokens are matched in full.

\subsection{Match Rules}
\label{sec:app-match}

Let $s$ be the normalized 256-token answer window. The four labels below are mutually exclusive and exhaustive; both/neither outputs are retained in the denominator rather than discarded.
\begin{itemize}[topsep=0pt,itemsep=1pt,parsep=0pt,leftmargin=*]
  \item \textbf{Parametric-only}: $s$ contains $a_\mathrm{par}$ as a substring, or an alias thereof, and does \textbf{not} contain $a_\mathrm{ctx}$.
  \item \textbf{Contextual-only}: $s$ contains $a_\mathrm{ctx}$ and does \textbf{not} contain $a_\mathrm{par}$.
  \item \textbf{Both}: $s$ contains both answers.
  \item \textbf{Neither}: $s$ contains neither answer, or the output is empty/refusal/off-topic.
\end{itemize}
Long answers are handled by substring search within the normalized answer window only; tokens after position 256 are not searched.
This operational metric measures answer-string source matching, not full semantic entailment; Appendix~\ref{app:complex-eval} therefore describes the benchmark-native semantic or constraint evaluators used for complex tasks.

\subsection{LLM-Judge Validation}
\label{app:judge-validation}
String matching permits deterministic, low-cost evaluation across the full alpha sweep. To audit its construct validity, we sample 100 OOD outputs per dataset. Given the prompt, both candidate answers, and the complete generation, a blinded LLM judge assigns one of the same four labels without seeing the string-match result. Table~\ref{tab:app-judge-agreement} compares the resulting four-way distributions; the close correspondence supports using string matching for the full sweeps.

\begin{table}[tbp]
  \centering
  \scriptsize
  \caption{Four-way label distributions (\%) from normalized string matching and a blinded LLM judge on 100 OOD outputs per dataset. The order is correct answer / conflicting distractor / both / neither.}
  \label{tab:app-judge-agreement}
  \begin{tabular}{@{}lcc@{}}
    \toprule
    Dataset & String matching & LLM judge \\
    \midrule
    ParaConflict & 57 / 9 / 34 / 0 & 59 / 10 / 30 / 1 \\
    MacNoise & 77 / 5 / 1 / 17 & 74 / 5 / 8 / 13 \\
    MemoTrap & 48 / 15 / 32 / 5 & 50 / 17 / 30 / 3 \\
    CounterFact & 53 / 20 / 23 / 4 & 54 / 24 / 18 / 4 \\
    \bottomrule
  \end{tabular}
\end{table}

\subsection{Complex-Task Evaluators}
\label{app:complex-eval}
For each model, all complex-task experiments use its direction learned on NQ-Swap, injected at that model's selected layer with coefficients fixed from NQ-Swap validation; no target-task direction, coefficient, or layer is estimated or relocalized. For each task, all coefficients are evaluated on the same deterministic sample subset, and we report the mean official per-item score times 100. These downstream scores assess input-grounded behavior: except for RGB, a correct output alone does not establish whether the unsteered error arose from factual priors, evidence localization, reasoning, or response generation.

\paragraph{Sampling and prompts.}
We evaluate at most 500 processed items per task, sampling without replacement with seed 43 when a pool exceeds 500 items; smaller pools are evaluated in full. The model receives only the task-supplied context, and credit is assigned against the contextual target or benchmark constraint. Thus, complex-task scores do not use the 256-token source-match protocol above.

\paragraph{RAG evaluators.}
RGB contains 100 counterfactual retrieval items. Up to three supplied passages support a \textit{fakeanswer}, and its official binary judge scores that contextual target rather than the world-factual answer. For CRAG, we sample 500 of 2,698 development items; ranked snippets are HTML-stripped, title-prefixed when available, concatenated in retrieval order, and capped at 12,000 characters. Its released evaluator handles exact, invalid, and semantic-answer cases.

\paragraph{Memory evaluators.}
For LoCoMo, 500 of 1,542 QA rows are sampled and the official category-aware multi-hop or token-F1 evaluator is averaged. LoCoMo-Plus evaluates all 401 cognitive items with its released binary constraint-consistency judge. LongMemEval evaluates its 500-item oracle split in full, caps dated relevant sessions at 12,000 characters, and uses its released answer-check judge conditioned on question type.

\paragraph{Rule-following evaluators.}
For IFEval, 500 of 541 prompts are sampled and its released strict evaluator gives one only when every instruction is satisfied. For FollowBench, 500 of 820 items are sampled; released template, rule-based, or discriminative evaluators give hard per-item satisfaction, and we report the macro average across six constraint categories.

\section{Implementation Details}
\label{app:implementation-details}

All experiments are inference-only; we do not fine-tune any model. 
We run each model on a single NVIDIA A100 GPU with 80GB memory using \texttt{bfloat16} precision and the Hugging Face \texttt{transformers} library. 
Unless otherwise specified, decoding is greedy with \texttt{do\_sample=False} and \texttt{max\_new\_tokens=200}. 
The base evaluation cohort consists of examples where the model answers the clean prompt correctly and is then evaluated under the conflict prompt, denoted as \texttt{stage1\_correct\_use\_wrong\_context}. 
When subsampling is required, we use random seed 42.

\begin{table}[tbp]
  \centering
  \footnotesize
  \setlength{\tabcolsep}{6pt}
  \renewcommand{\arraystretch}{1.08}
  \caption{Model checkpoints and selected intervention layers.}
  \label{tab:app-model-layer}
  \begin{tabular}{@{}l c c c@{}}
    \toprule
    Model & Size & Local key & Layer \\
    \midrule
    Llama-3.1-8B-Instruct & 8B & \texttt{llama3-8b-it} & 10 \\
    Gemma-2-9B-IT       & 9B & \texttt{gemma2-9b-it} & 16 \\
    Qwen3-8B            & 8B & \texttt{qwen3-8b} & 21 \\
    Mistral-7B-Instruct-v0.3 & 7B & \texttt{mistral-7b-instruct-v0.3} & 16 \\
    \bottomrule
  \end{tabular}
\end{table}

\begin{table}[tbp]
  \centering
  \footnotesize
  \setlength{\tabcolsep}{7pt}
  \renewcommand{\arraystretch}{1.08}
  \caption{Inference and tokenization settings.}
  \label{tab:app-inference-settings}
  \begin{tabular}{@{}l l@{}}
    \toprule
    Setting & Value \\
    \midrule
    Framework & Hugging Face \texttt{transformers} \\
    Precision & \texttt{bfloat16} \\
    Hardware & 1$\times$ NVIDIA A100 80GB \\
    Attention implementation & \texttt{eager} \\
    Decoding & Greedy, \texttt{do\_sample=False} \\
    \texttt{max\_new\_tokens} & 200 \\
    Tokenizer \texttt{max\_length} & 8192 \\
    Padding side & Left padding \\
    \texttt{pad\_token} & \texttt{eos\_token} if unset \\
    Batch size & 128 \\
    Batch size for NQ-Swap & 64 \\
    GPUs per job & 1, set by \texttt{CUDA\_VISIBLE\_DEVICES} \\
    Random seed & 42 when subsampling is used \\
    \bottomrule
  \end{tabular}
\end{table}
\section{Steering-Strength Sensitivity Analysis}
\label{app:main-alpha-stability}
For the Llama-3.1-8B-Instruct direction transferred from NQ-Swap, we report the selected coefficient and the two nearest evaluated values on either side when available. Table~\ref{tab:app-main-alpha-stability} gives parametric-answer rate for positive steering and distractor/context-answer rate for negative steering. No target-side coefficient is reselected.

\begin{table*}[tbp]
  \centering
  \scriptsize
  \caption{Cross-dataset operating-point stability for Llama-3.1-8B-Instruct. Each cell gives $\alpha$: held-out source-match rate (\%) followed by the number of valid Stage-2 generations in parentheses. The selected transfer coefficient is bold; ``--'' indicates that a farther neighboring grid point was not evaluated.}
  \label{tab:app-main-alpha-stability}
  \begin{tabular}{@{}lcccc@{}}
    \toprule
    Setting & ParaConflict & MacNoise & MemoTrap & CounterFact \\
    \midrule
    \multicolumn{5}{@{}l@{}}{\textit{Positive direction: parametric-answer rate}} \\
    $\alpha=0$ & 15.9 (1711) & 46.7 (1047) & 67.0 (624) & 36.1 (2160) \\
    Lower 2 & 1.2: 82.3 (1701) & 1.2: 71.1 (849) & 2.4: 83.7 (509) & 2.6: 43.7 (1854) \\
    Lower 1 & 1.4: 84.7 (1697) & 1.4: 72.1 (795) & 2.6: 82.7 (490) & 2.8: 43.6 (1803) \\
    Selected & \textbf{1.6: 85.6 (1687)} & \textbf{1.6: 72.4 (749)} & \textbf{2.8: 85.6 (465)} & \textbf{3.0: 48.1 (1565)} \\
    Higher 1 & 1.8: 84.3 (1674) & 1.8: 70.4 (689) & 3.0: 83.1 (450) & -- \\
    Higher 2 & 2.0: 81.6 (1662) & 2.0: 62.3 (159) & -- & -- \\
    \midrule
    \multicolumn{5}{@{}l@{}}{\textit{Negative direction: distractor/context-answer rate}} \\
    $\alpha=0$ & 81.8 (1711) & 27.3 (1047) & 30.1 (624) & 60.8 (2160) \\
    Lower 2 & $-1.8$: 98.3 (1264) & $-1.8$: 23.4 (124) & -- & $-1.4$: 61.6 (1914) \\
    Lower 1 & $-1.6$: 98.6 (1280) & $-1.6$: 21.1 (147) & $-1.8$: 57.6 (580) & $-1.2$: 62.9 (1957) \\
    Selected & \textbf{$-1.4$: 98.8 (1292)} & \textbf{$-1.4$: 43.8 (706)} & \textbf{$-2.0$: 58.0 (571)} & \textbf{$-1.0$: 63.0 (2005)} \\
    Higher 1 & $-1.2$: 96.8 (1528) & $-1.2$: 43.1 (796) & -- & $-0.8$: 62.8 (2038) \\
    Higher 2 & $-1.0$: 96.9 (1578) & $-1.0$: 42.2 (855) & -- & $-0.6$: 62.8 (2065) \\
    \bottomrule
  \end{tabular}
\end{table*}

\section{Scaling to Larger Models}
\label{sec:large-models}

\begin{figure*}[tbp]
  \centering
  \includegraphics[width=\textwidth]{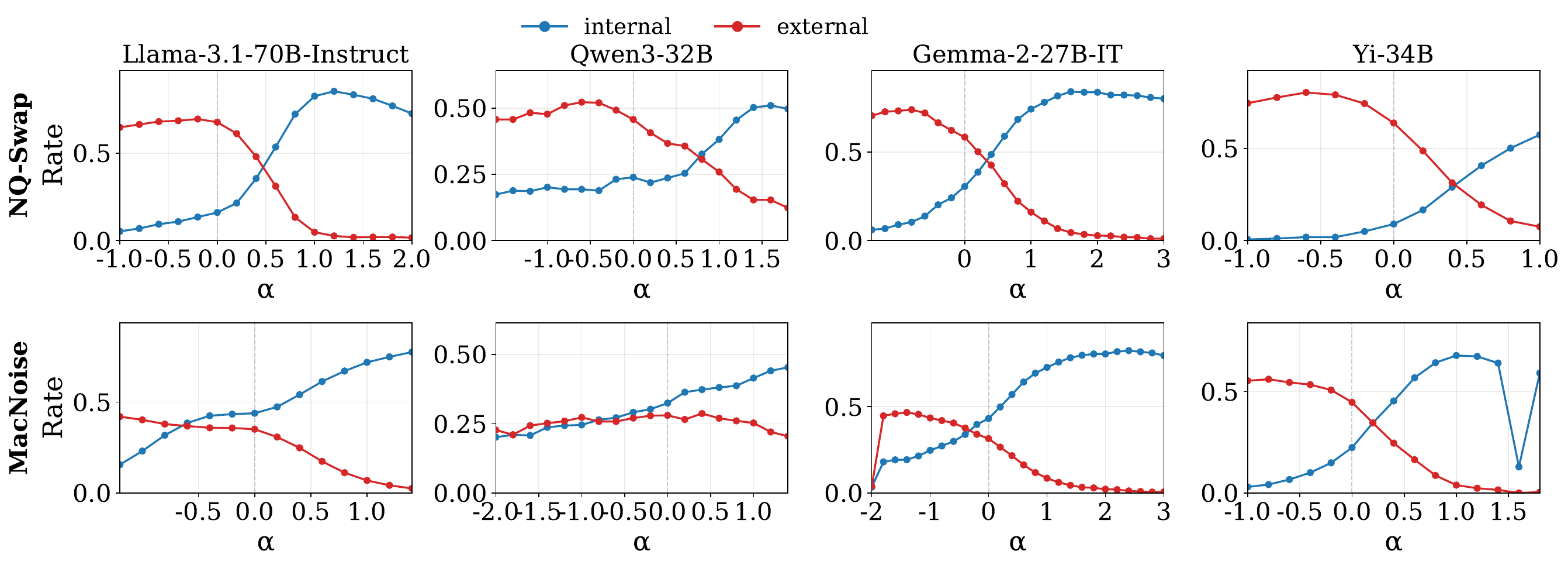}
  \caption{Source-choice trajectories on larger models. 
  Blue curves denote internal, or parametric, source-choice rates, while red curves denote external, or contextual, source-choice rates. 
  Across {NQ-Swap} and {MacNoise}, increasing $\alpha$ generally shifts the model from external reliance toward internal reliance, showing that the steering direction remains effective for larger models.}
  \label{fig:large-curves}
\end{figure*}

We further examine whether the same one-dimensional steering behavior holds for larger instruction-tuned models. 
Figure~\ref{fig:large-curves} reports source-choice rates on {NQ-Swap} and {MacNoise} for four larger instruction-tuned models. 
Across most models and datasets, increasing $\alpha$ raises the internal, or parametric, source-choice rate while reducing the external, or contextual, source-choice rate. 
This consistent trend indicates that the learned direction remains effective beyond the smaller models used in the main experiments, suggesting that knowledge selection can be continuously modulated along a similar residual-stream direction even at larger model scales.

\section{Instance-Level Changes and Failure Analysis}
\label{sec:app-steering-audit}

This audit classifies baseline context-conflict errors that an available positive-$\alpha$ sweep changed into an officially correct output. Here, knowledge conflict is used broadly: a model default (factual prior, continuation, event ordering, or output habit) can conflict with information or constraints supplied by the context. The alpha shown for each example is the first observed correct positive-alpha output, not an item-level selection rule. ``Parametric knowledge'' is an inference from the baseline text, not a claim about an unobserved model state. We use three mechanism labels: \textbf{P} for an explicit override of supplied evidence by a real-world or general-knowledge correction; \textbf{L} for missed, misordered, or underspecified supplied context; and \textbf{I} for a requested format or constraint that the context requires.

\subsection{Controlled Factual-Conflict Instance-Level Changes}
\label{sec:app-four-model-cases}

\newcommand{\casetitle}[1]{%
  \par\addvspace{0.6em}
  \noindent\textbf{Example #1.}\par
  \vspace{1pt}
}

Each trajectory contains an unsteered output, an intermediate output containing both answer strings, and an output selecting the requested source. Llama, Qwen, and Gemma cases use saved NQ-Swap test alpha grids. Mistral retained only three final NQ-Swap coefficients, so its cases use the saved complete CounterFact within-relation grid. Outputs are answer-bearing excerpts; \textsc{both} denotes a string-match label, not a claim that the full generation provides a coherent answer.

\newcommand{\trajrow}[4]{#1 & \textit{Initial}: #2\newline\textit{Intermediate}: #3\newline\textit{Target}: #4 \\
}
\newcommand{\trajectorytable}[2]{%
\begin{table*}[tbp]
\centering\small
\caption{Instance-level continuous trajectories for #1. P and C denote the requested parametric and contextual sources.}
\label{#2}
\setlength{\tabcolsep}{5pt}
\renewcommand{\arraystretch}{1.08}
\begin{tabular}{@{}p{.28\textwidth}p{.66\textwidth}@{}}
\toprule
\textbf{Question} & \textbf{Generation trajectory} \\
\midrule
}
\newcommand{\trajectoryend}{\bottomrule\end{tabular}\end{table*}}

\paragraph{Llama-3.1-8B-Instruct.}
Table~\ref{tab:app-cases-llama} gives three parametric and three contextual trajectories for the main model, illustrating the transition from an initial source choice through a mixed string-match output to the requested source.
\trajectorytable{Llama-3.1-8B-Instruct}{tab:app-cases-llama}
\rowcolor{parUnsteer}\multicolumn{2}{@{}l@{}}{\textbf{P: requested parametric source}} \\
\trajrow{When did the first Ninja Turtles come out?}{1999.}{The first comic was published in 1984, not 1999.}{1984.}
\trajrow{Who designed the Earth Day flag in 1969?}{Andreas Vesalius.}{Vesalius \ldots likely John McConnell.}{John McConnell.}
\trajrow{Who was in the \emph{Don't Worry, Be Happy} video?}{CeCe Drake.}{CeCe Drake. Bobby McFerrin \ldots}{Bobby McFerrin.}
\midrule
\rowcolor{ctxUnsteer}\multicolumn{2}{@{}l@{}}{\textbf{C: requested contextual source}} \\
\trajrow{When was \emph{Edward Tulane} published?}{The novel is from 2006.}{15th century \ldots a 21st-century novel.}{15th century.}
\trajrow{How many episodes is season 4 of \emph{The Flash}?}{23 episodes.}{The correct answer should be 23, not 360.}{360 episodes.}
\trajrow{When was the \$1,000 bill discontinued?}{1969.}{2017 \ldots discontinued in 1969.}{2017.}
\trajectoryend

\paragraph{Qwen3-8B.}
Table~\ref{tab:app-cases-qwen} repeats the same trajectory audit for Qwen3-8B, showing that both source directions can be expressed through intermediate mixed outputs.
\trajectorytable{Qwen3-8B}{tab:app-cases-qwen}
\rowcolor{parUnsteer}\multicolumn{2}{@{}l@{}}{\textbf{P: requested parametric source}} \\
\trajrow{When did \emph{The Karate Kid} come out?}{The text says the remake was in 1835.}{The original was 1984, and the remake was 2010.}{The remake was released in 2010.}
\trajrow{How many grams of alcohol are in one beer?}{The text says a standard drink contains 1612 grams.}{The correct value should be 14 grams, not 1612.}{A standard drink contains about 14 grams.}
\trajrow{Industrial city on Germany's Rhine--Herne Canal?}{The canal connects near Hawaii.}{It connects near Hawaii \ldots Duisburg.}{It connects the Rhine River near Duisburg.}
\midrule
\rowcolor{ctxUnsteer}\multicolumn{2}{@{}l@{}}{\textbf{C: requested contextual source}} \\
\trajrow{Total presidential electoral votes?}{The total is 538.}{The total is 538, not 201.}{201.}
\trajrow{When was \emph{Harry Potter and the Philosopher's Stone} published?}{It was first published in 1997.}{The answer is 11 November 1918; the correct date should be 1997.}{The text says it was first published in 11 November 1918.}
\trajrow{When was the first \emph{Star Wars} film released?}{The first film was released in 1977.}{The trilogy was released from 1889 \ldots the first film was 1977.}{The original trilogy was released between 1889 and 1983.}
\trajectoryend

\paragraph{Gemma-2-9B-IT.}
Table~\ref{tab:app-cases-gemma} reports the corresponding Gemma trajectories; the examples include both clean source switches and cases where the model retains a competing answer in the intermediate generation.
\trajectorytable{Gemma-2-9B-IT}{tab:app-cases-gemma}
\rowcolor{parUnsteer}\multicolumn{2}{@{}l@{}}{\textbf{P: requested parametric source}} \\
\trajrow{Who overthrew the Mongols and established the Ming dynasty?}{Cliff Williams.}{Cliff Williams (Zhu Yuanzhang).}{Zhu Yuanzhang.}
\trajrow{When was the first Ford F-150 made?}{1994.}{1994. This is incorrect; it was made in 1975.}{1975.}
\trajrow{When did the Right to Buy scheme start?}{March 31.}{March 31, 1980.}{The scheme was introduced in March 1980.}
\midrule
\rowcolor{ctxUnsteer}\multicolumn{2}{@{}l@{}}{\textbf{C: requested contextual source}} \\
\trajrow{Where did the \emph{Titanic} sink?}{The Titanic sank in the North Atlantic Ocean.}{It sank in Saint Petersburg in the North Atlantic Ocean.}{The Titanic sank in Saint Petersburg.}
\trajrow{Nuclear plant that blew up in Russia?}{Chernobyl.}{Chernobyl \ldots the text does not describe a Colorado disaster.}{Colorado Nuclear Power Plant.}
\trajrow{Which country joined Mexico and the US to form NAFTA?}{Canada.}{Brooklyn. This is incorrect; the answer is Canada.}{Brooklyn.}
\trajectoryend

\paragraph{Mistral-7B-Instruct-v0.3.}
Table~\ref{tab:app-cases-mistral} reports Mistral's within-relation CounterFact trajectories, using the saved CounterFact grid because a complete NQ-Swap alpha sweep was unavailable for this model.
\trajectorytable{Mistral-7B-Instruct-v0.3 on CounterFact}{tab:app-cases-mistral}
\rowcolor{parUnsteer}\multicolumn{2}{@{}l@{}}{\textbf{P: requested parametric source}} \\
\trajrow{Abd Allah ibn al-Zubayr's religion?}{Buddhism.}{Buddhism. Abd Allah ibn al-Zubayr \ldots}{Islam.}
\trajrow{Kasey Keller is a professional?}{Football player, specifically a goalkeeper.}{Footballer \ldots one of the greatest in American soccer.}{Soccer player who played as a goalkeeper.}
\trajrow{Manolache Costache Epureanu citizenship?}{Cambodia.}{Romania and Cambodia.}{Romania.}
\midrule
\rowcolor{ctxUnsteer}\multicolumn{2}{@{}l@{}}{\textbf{C: requested contextual source}} \\
\trajrow{What sport does Corey Perry play?}{Ice hockey.}{Corey Perry is an ice hockey player, not a soccer player.}{Soccer.}
\trajrow{BMW M52, developed by?}{BMW, a straight-6 engine.}{Sony, a 2.5-liter inline-six engine used in BMW automobiles.}{Sony, produced by Sony.}
\trajrow{Dokka Umarov's religion?}{Islam.}{Umarov is a follower of Judaism. Islam.}{Judaism.}
\trajectoryend

\subsection{Complex-Task Instance-Level Changes}
\label{sec:app-audit-cases}

The audit is reported in Table~\ref{tab:app-complex-cases} and Table~\ref{tab:app-complex-cases-part2}, with one row per representative item. The labels distinguish explicit parametric overrides (P), localization or reasoning over supplied context (L), and instruction constraints (I). RGB provides synthetic counterfactual conflicts; CRAG mixes evidence selection with temporal, arithmetic, and comparison cases; the memory benchmarks distribute the relevant fact across earlier turns; and IFEval/FollowBench test contextual constraints on output form.

\subsubsection{Gain stratification and limitations}
\label{sec:app-audit-stratification}

\begin{table}[tbp]
\centering
\footnotesize
\caption{Audit-supported interpretation of the reported Llama-3.1-8B-Instruct gains. ``N/A'' denotes an unobserved stratum, not zero gain.}
\label{tab:app-audit-stratification}
\begin{tabular}{@{}p{0.20\linewidth}p{0.25\linewidth}p{0.20\linewidth}p{0.25\linewidth}@{}}
\toprule
Dataset & Conflict mechanism in audit & Observed gain & Interpretation \\
\midrule
RGB & Dominant P by construction & $+20.0$ (62.0$\to$82.0) & Synthetic counterfactual adherence; no matched non-conflict cohort \\
CRAG & Mixed P/L candidates & $+2.6$ (23.7$\to$25.8) & Both kinds may occur; audit is illustrative, not random \\
LoCoMo & Distributed-context retrieval (L) & $+4.6$ (8.5$\to$13.2) & Dialogue-fact localization and temporal grounding \\
LoCoMo-Plus & Earlier-turn grounding (L) & $+5.0$ (32.7$\to$36.7) & Connecting a later request to an earlier event \\
LongMemEval & Memory retrieval and ordering (L) & $+3.0$ (43.2$\to$45.4) & Selecting the relevant personal-memory fact \\
IFEval / FollowBench & Instruction constraints (I) & $+3.0$ / $+0.4$ & Instruction-following controls over output length and format \\
\bottomrule
\end{tabular}
\end{table}

The current sweeps support a broad context-conflict account spanning factual evidence, distributed memory, and instruction constraints, while the mechanism-level comparison remains qualitative. A stronger test would sample 50--100 CRAG baseline errors before inspecting steered outputs, independently label evidence spans and confidence, freeze $\alpha$ on validation data, and report per-mechanism test gains with bootstrap confidence intervals.

\begin{table*}[tbp]
\centering
\scriptsize
\caption{Representative complex-task cases from the available positive-alpha sweeps (Part I). Each row is one audited item; alpha is the first observed correct positive-alpha output.}
\label{tab:app-complex-cases}
\begin{tabular}{@{}p{0.13\linewidth}p{0.39\linewidth}p{0.25\linewidth}cp{0.12\linewidth}@{}}
\toprule
\textbf{Family} & \textbf{Question / case} & \textbf{Baseline $\rightarrow$ steered} & $\alpha$ & \textbf{Label} \\
\midrule
RGB & 2013 Wimbledon men's singles & Andy Murray $\rightarrow$ Dominic Thiem & 1.5 & P \\
RGB & 2017 French presidential election & Emmanuel Macron $\rightarrow$ Marine Le Pen & 2 & P \\
RGB & 2020 US presidential election & Joe Biden $\rightarrow$ Donald Trump & 1.5 & P \\
RGB & GPT-3 release date & June 2020 $\rightarrow$ June 2010 & 1.5 & P \\
RGB & 2018 World Cup final & France $\rightarrow$ Germany & 1.5 & P \\
CRAG & Current US president's age & 78 at inauguration $\rightarrow$ 81 currently & 1.5 & P/L \\
CRAG & Best animated-feature Oscar in 2018 & Spider-Verse $\rightarrow$ Coco & 1.5 & P/L \\
CRAG & Film released first & The Notebook $\rightarrow$ A Walk to Remember & 2 & L \\
CRAG & Lincoln's age at inauguration & 51 $\rightarrow$ 52 & 1.5 & P/L \\
CRAG & Largest country bordering Pacific & rejects Russia $\rightarrow$ Russia & 2 & P/L \\
\bottomrule
\end{tabular}
\end{table*}

\begin{table*}[tbp]
\centering
\scriptsize
\caption{Representative complex-task cases (Part II).}
\label{tab:app-complex-cases-part2}
\begin{tabular}{@{}p{0.13\linewidth}p{0.39\linewidth}p{0.25\linewidth}cp{0.12\linewidth}@{}}
\toprule
\textbf{Family} & \textbf{Question / case} & \textbf{Baseline $\rightarrow$ steered} & $\alpha$ & \textbf{Label} \\
\midrule
LoCoMo & Jolene's diving location & vague dive spot $\rightarrow$ Phuket & 1.5 & L \\
LoCoMo & Can Dave work with engines? & no evidence $\rightarrow$ yes & 0.5 & L \\
LoCoMo & Dave's hobby in October 2023 & absent $\rightarrow$ photography & 0.5 & L \\
LoCoMo-Plus & Learning to say no, then volunteering & generic lesson $\rightarrow$ connects volunteering & 1 & L \\
LoCoMo-Plus & Identity theft and shredding receipts & generic lesson $\rightarrow$ names identity theft & 1 & L \\
LoCoMo-Plus & Peanut allergy and restaurant cancellation & generic advice $\rightarrow$ cancellation/anxiety & 1.5 & L \\
LoCoMo-Plus & Mother's missed cancer diagnosis & generic reassurance $\rightarrow$ names diagnosis/checkups & 1.5 & L \\
LoCoMo-Plus & Plant-based meals and craving cheese & generic follow-up $\rightarrow$ contrasts prior decision & 0.5 & L \\
LongMemEval & Admon's Sunday social-media shift & night shift $\rightarrow$ day shift & 1.5 & L \\
LongMemEval & First attended event & none $\rightarrow$ Python webinar & 1.5 & L \\
LongMemEval & Platform with most new followers & refuses comparison $\rightarrow$ TikTok & 0.5 & L \\
LongMemEval & Location of Ethereal Dreams & refuses current location $\rightarrow$ bedroom & 1.5 & L \\
LongMemEval & Device acquired first & Dell XPS $\rightarrow$ Samsung Galaxy S22 & 1.5 & L \\
IFEval & Four-section clown resume & Markdown headings $\rightarrow$ Section 1--4 & 1.5 & I \\
IFEval & Worms versus dead cow & wrong separators $\rightarrow$ required divider/blocks & 1.5 & I \\
IFEval & Four-bullet home-theater rubric & prose/mixed bullets $\rightarrow$ four bullets & 2 & I \\
IFEval & Bremen sweatshirt proposal & headings $\rightarrow$ highlighted labels & 2 & I \\
IFEval & JSON travel response & banned words remain $\rightarrow$ compliant destinations & 1.5 & I \\
FollowBench & Short story title/alliterative introduction & non-compliant $\rightarrow$ alliterative format & 0.5 & I \\
FollowBench & Time advice with mind map & implicit $\rightarrow$ explicit mind map & 1.5 & I \\
FollowBench & Broadway actors, concise alliteration & length/style violations $\rightarrow$ compliant bullets & 0.5 & I \\
FollowBench & Cube-Earth gravity as three haikus & wrong structure $\rightarrow$ three titled haikus & 1.5 & I \\
FollowBench & TCP/IP one protocol per bullet & prose/mixed layout $\rightarrow$ one per layer & 2 & I \\
\bottomrule
\end{tabular}
\end{table*}

\subsection{Failure Analysis}
\label{app:error-analysis}

\newcommand{\errorcaseinfo}[3]{%
\noindent
\begin{tabular}{@{}p{0.22\linewidth}p{0.70\linewidth}@{}}
\textit{Question:} & #1 \\
\textit{Context:} & #2 \\
\textit{Parameter:} & #3 \\
\end{tabular}
\vspace{2pt}
}

\noindent
We analyze representative failure cases of steering. 
Here, \textsc{parameter} denotes the gold or internally stored answer, \textsc{context} denotes the distractor supported by the conflict prompt, \textsc{both} denotes outputs containing both answers, and \textsc{other} denotes outputs containing neither.

\subsubsection{Failure to Reach the Target Source}
\label{app:error-f5}

In this category, steering does not move the first answer entity to the intended knowledge source. 
For positive steering, the output remains contextual; for negative steering, the output remains parametric.

\casetitle{1}
\errorcaseinfo
{George V Coast is a part of the continent of}
{Africa}
{Antarctica}

\begin{center}
\footnotesize
\setlength{\tabcolsep}{3pt}
\renewcommand{\arraystretch}{1.08}
\begin{tabular}{@{}c p{0.56\linewidth} c@{}}
\toprule
$\alpha$ & Output & Label \\
\midrule
$0$
& ``Africa.''
& \textsc{context} \\
$+0.2$
& ``Africa.''
& \textsc{context} \\
$+1.0$
& ``Africa.''
& \textsc{context} \\
$+2.4$
& ``Africa, and it is a region that is rich in natural resources.''
& \textsc{context} \\
$+3.0$
& ``Africa, and it is a region of great scientific interest due to its unique geology and geography.''
& \textsc{context} \\
\bottomrule
\end{tabular}
\end{center}

\noindent
The conflict prompt anchors the first answer to \textit{Africa}; even strong positive steering fails to recover the parametric answer \textit{Antarctica}.

\casetitle{2}
\errorcaseinfo
{what country shares borders with both belarus and romania}
{Kentucky}
{Ukraine}

\begin{center}
\footnotesize
\setlength{\tabcolsep}{3pt}
\renewcommand{\arraystretch}{1.08}
\begin{tabular}{@{}c p{0.56\linewidth} c@{}}
\toprule
$\alpha$ & Output & Label \\
\midrule
$0$
& ``Ukraine Explanation: The text states that Ukraine shares borders with Belarus and Romania.''
& \textsc{parameter} \\
$-0.1$
& ``Ukraine Explanation: The text states that Ukraine shares borders with Belarus and Romania.''
& \textsc{parameter} \\
$-0.3$
& ``Ukraine Note: The question is based on the text, which is about the borders of Kentucky.''
& \textsc{both} \\
$-0.6$
& ``Ukraine.''
& \textsc{parameter} \\
$-1.0$
& ``Ukraine.''
& \textsc{parameter} \\
\bottomrule
\end{tabular}
\end{center}

\noindent
Negative steering briefly introduces the contextual entity in a mixed explanation, but the first answer remains \textit{Ukraine}; the model never switches to the target contextual answer \textit{Kentucky}.

\subsubsection{Off-Topic Best Output}
\label{app:error-f4}

In this category, the best steered output is neither parametric nor contextual. 
Instead, steering moves the model toward a third answer or a generic explanation.

\casetitle{1}
\errorcaseinfo
{when was the last time the military drafted}
{7th century}
{1973}

\begin{center}
\footnotesize
\setlength{\tabcolsep}{3pt}
\renewcommand{\arraystretch}{1.08}
\begin{tabular}{@{}c p{0.56\linewidth} c@{}}
\toprule
$\alpha$ & Output & Label \\
\midrule
$0$
& ``The last time the military drafted was during the Vietnam War.''
& \textsc{other} \\
$+0.2$
& ``The last time the United States military drafted was during the Vietnam War.''
& \textsc{other} \\
$+0.8$
& ``The last time the United States military drafted was during the Vietnam War, which ended in 1975.''
& \textsc{other} \\
$+1.4$
& ``The last time the United States military drafted was during the Vietnam War, which ended in 1975.''
& \textsc{other} \\

\bottomrule
\end{tabular}
\end{center}

\noindent
Positive steering does not recover the exact parametric answer \textit{1973}; instead, the model settles on a related but off-target answer, \textit{Vietnam War} or \textit{1975}.

\casetitle{2}
\errorcaseinfo
{what college does everyone in gossip girl go to}
{St Pancras International}
{New York University, Columbia University}

\begin{center}
\footnotesize
\setlength{\tabcolsep}{3pt}
\renewcommand{\arraystretch}{1.08}
\begin{tabular}{@{}c p{0.56\linewidth} c@{}}
\toprule
$\alpha$ & Output & Label \\
\midrule
$0$
& ``Yale University The best answer is Yale.''
& \textsc{other} \\
$-0.1$
& ``Constance Billard and St.''
& \textsc{other} \\
$-0.3$
& ``Constance Note: Constance is a private school in the Gossip Girl series, not a college.''
& \textsc{other} \\

$-1.0$
& ``Constance Billard and St.''
& \textsc{other} \\
\bottomrule
\end{tabular}
\end{center}

\noindent
Negative steering does not produce the contextual answer \textit{St Pancras International}; instead, the model falls back to other salient entities from the show, such as \textit{Yale} or \textit{Constance}.

\subsubsection{Mixed Output Remains Best}
\label{app:error-f2}

In this category, steering partially changes the output but fails to suppress the competing source. 
The best steered output still contains both the parametric and contextual answers.

\casetitle{1}
\errorcaseinfo
{where did the titanic sink at what ocean}
{Saint Petersburg}
{North Atlantic Ocean}

\begin{center}
\footnotesize
\setlength{\tabcolsep}{3pt}
\renewcommand{\arraystretch}{1.08}
\begin{tabular}{@{}c p{0.56\linewidth} c@{}}
\toprule
$\alpha$ & Output & Label \\
\midrule
$0$
& ``The Titanic sank in the Saint Petersburg in the North Atlantic Ocean.''
& \textsc{both} \\
$-0.1$
& ``The titanic sank at the Saint Petersburg in the North Atlantic Ocean.''
& \textsc{both} \\
$-0.5$
& ``The titanic sank at the Saint Petersburg in the North Atlantic Ocean.''
& \textsc{both} \\
$-0.9$
& ``The titanic sank at the Saint Petersburg in the North Atlantic Ocean.''
& \textsc{both} \\
$-1.0$
& ``the titanic sank at the saint petersburg in the saint petersburg is not a ocean but \dots\ in the north atlantic ocean \dots''
& \textsc{both} \\
\bottomrule
\end{tabular}
\end{center}

\noindent
The contextual distractor \textit{Saint Petersburg} remains present throughout negative steering, while the parametric answer \textit{North Atlantic Ocean} is not removed.
\casetitle{2}
\errorcaseinfo
{who sang nice day for a white wedding}
{Dido}
{Billy Idol}

\begin{center}
\footnotesize
\setlength{\tabcolsep}{3pt}
\renewcommand{\arraystretch}{1.08}
\begin{tabular}{@{}c p{0.56\linewidth} c@{}}
\toprule
$\alpha$ & Output & Label \\
\midrule
$0$
& ``Dido Note: The song is actually called `White Wedding' and not `Nice Day for a White Wedding'.''
& \textsc{context} \\
$+0.2$
& ``Dido Note: The song is actually called `White Wedding' and not `Nice Day for a White Wedding'.''
& \textsc{context} \\
$+0.4$
& ``Dido Note: The song `White Wedding' is actually by Billy Idol, not Dido.''
& \textsc{both} \\
$+0.8$
& ``Dido Note: `White Wedding' is a song by Billy Idol, although the context mentions Dido.''
& \textsc{both} \\
$+1.2$
& ``Dido is mentioned in the context, but the song `White Wedding' is by Billy Idol.''
& \textsc{both} \\
$+2.0$
& ``The context points to Dido, while the parametric answer is Billy Idol.''
& \textsc{both} \\
\bottomrule
\end{tabular}
\end{center}

\noindent
Positive steering briefly introduces the parametric answer \textit{Billy Idol}, but the output still begins with and retains the contextual distractor \textit{Dido}. 
Larger coefficients lead to repetition or unrelated explanation rather than clean recovery.

\section{Prompt Templates}
\label{app:prompt-templates}

This appendix records the raw prompts used in the experiments. Each raw prompt is passed as a single user message through the model tokenizer's chat template when applicable; Qwen3 uses \texttt{enable\_thinking=False}. There is no shared system prompt or generation suffix. A run without a chat template uses the raw prompt directly.

\paragraph{General format.}
\noindent\textbf{General: Clean Prompt}
\begin{quote}
\small\ttfamily
Question: \{question\}\\
Answer:
\end{quote}

\noindent\textbf{General: Conflict Prompt}
\begin{quote}
\small\ttfamily
\{wrong\_context\}\\
Question: \{question\}\\
Answer:
\end{quote}

\paragraph{NQ-Swap context controls.}
For a question $Q$, the A/B/C/D control prompts are respectively: no context, the original context, the substituted context, and the original context from the next NQ-Swap row. B, C, and D share the context-plus-question form above. The four directions are mean(A)--mean(C), mean(B)--mean(C), mean(A)--mean(B), and mean(A)--mean(D).

\noindent\textbf{A: No context}\
\begin{quote}\small\ttfamily Question: \{Q\}\\ Answer:\end{quote}
\noindent\textbf{B/C/D: Aligned, conflicting, or irrelevant context}\
\begin{quote}\small\ttfamily \{CONTEXT\}\\[2pt] Question: \{Q\}\\ Answer:\end{quote}

\paragraph{NQ-Swap.}
The clean prompt contains only the question, while the conflict prompt prepends the substituted context \texttt{sub\_context}. 
In preprocessing, the original \texttt{<P>} and \texttt{</P>} markers are replaced with spaces.

\noindent\textbf{NQ-Swap: Clean Prompt}
\begin{quote}
\small\ttfamily
Question: how many episodes are in chicago fire season 4\\
Answer:
\end{quote}

\noindent\textbf{NQ-Swap: Conflict Prompt}
\begin{quote}
\small\ttfamily
The fourth season of Chicago Fire, an American drama television series with executive producer Dick Wolf, \dots\ premiered on October 13, 2015 and concluded on May 17, 2016. The season contained 775 episodes.\\
Question: how many episodes are in chicago fire season 4\\
Answer:
\end{quote}

\paragraph{MacNoise.}
The conflict prompt uses the perturbed context field \texttt{ctxs[i].text\_pert}, where $i$ is the first index satisfying \texttt{pert\_lbls\_100 == 1}.

\noindent\textbf{MacNoise: Clean Prompt}
\begin{quote}
\small\ttfamily
Question: who got the first nobel prize in physics\\
Answer:
\end{quote}

\noindent\textbf{MacNoise: Conflict Prompt}
\begin{quote}
\small\ttfamily
title: Wilhelm Röntgen context: Wilhelm Röntgen Cindy Bond (; ; 27 March 1845 -- 10 February 1923) was a German mechanical engineer and physicist, \dots\ an achievement that earned him the first Nobel Prize in Physics in 1901 \dots\\
Question: who got the first nobel prize in physics\\
Answer:
\end{quote}

\paragraph{ParaConflict.}
The clean prompt is the supplied \,\texttt{CLEAN\_PROMPT}. The conflict prompt is the supplied \textit{Substitution Conflict} or \textit{Coherent Conflict} field; the conflict-style setting selects substitution or coherent, and \,\texttt{auto} prefers coherent. An aligned version replaces the \textit{Distracted Token} with the first gold \textit{Answer}, preserving the conflict-template structure; failed replacements fall back to the clean prompt.

\noindent\textbf{ParaConflict: Clean Prompt}
\begin{quote}
\small\ttfamily
Conor McGregor plays the sport of
\end{quote}

\noindent\textbf{ParaConflict: Coherent Conflict Prompt}
\begin{quote}
\small\ttfamily
Conor McGregor plays the sport of basketball. Recognized by peers and fans alike, Conor McGregor's journey in basketball has been highlighted in various sports publications, interviews, and athlete profiles. \dots\\
Question: What sport does Conor McGregor play?\\
Answer: Conor McGregor plays the sport of
\end{quote}

\paragraph{CounterFact.}
The clean prompt is the original factual prefix. 
The conflict prompt concatenates \texttt{paraphrase\_prefix}, the substituted target \texttt{target\_new}, and the original base query \texttt{base\_q}.

\noindent\textbf{CounterFact: Clean Prompt}
\begin{quote}
\small\ttfamily
Mahmoud Fawzi has a citizenship from
\end{quote}

\noindent\textbf{CounterFact: Conflict Prompt}
\begin{quote}
\small\ttfamily
List of administrators\\
References\\
* Mahmoud Fawzi, who is a citizen of Germany.\\
Mahmoud Fawzi has a citizenship from
\end{quote}

\paragraph{MemoTrap.}
The clean prompt uses the cleaned completion prefix, while the conflict prompt keeps the original instruction, creating a conflict between the memorized phrase continuation and the instruction-imposed ending word.

\noindent\textbf{MemoTrap: Clean Prompt}
\begin{quote}
\small\ttfamily
Write a quote : Absence makes the heart grow
\end{quote}

\noindent\textbf{MemoTrap: Conflict Prompt}
\begin{quote}
\small\ttfamily
Write a quote that ends in the word ``heavy'': Absence makes the heart grow
\end{quote}

\paragraph{RGB.}
We concatenate at most three passages from \texttt{positive\_wrong}, separated by blank lines and truncated to 2,500 characters. The context-side target is \texttt{fakeanswer}.
\begin{quote}\small\ttfamily
\{POSITIVE\_WRONG\_1\}\\[2pt]
\{POSITIVE\_WRONG\_2\}\\[2pt]
\{POSITIVE\_WRONG\_3\}\\[4pt]
Question: \{QUERY\}\\
Answer:
\end{quote}

\paragraph{CRAG.}
Retrieved \texttt{search\_results} remain in ranking order. Each \texttt{page\_snippet} is HTML-stripped, whitespace-normalized, and capped at 4,000 characters; when available, \texttt{page\_name} precedes the snippet. The packed context is capped at 12,000 characters.
\begin{quote}\small\ttfamily
\{PAGE\_NAME\_1\}\\
\{CLEAN\_SNIPPET\_1\}\\[2pt]
\{PAGE\_NAME\_2\}\\
\{CLEAN\_SNIPPET\_2\}\\[4pt]
Question: \{QUERY\}\\
Answer:
\end{quote}

\paragraph{LoCoMo.}
History is retained as complete sessions, selected backward within a 12,000-character budget and restored to chronological order.
\begin{quote}\small\ttfamily
Conversation:\\[2pt]
\#\#\# Session \{N\} (\{SESSION\_DATE\})\\
\{SPEAKER\_1\}: \{TEXT\_1\}\\
\{SPEAKER\_2\}: \{TEXT\_2\}\\[4pt]
Question: \{QUESTION\}\\
Answer:
\end{quote}
For temporal items, the question appends \texttt{Use DATE of CONVERSATION to answer with an approximate date.}; adversarial items append the two supplied answer options, including ``Not mentioned in the conversation.''

\paragraph{LoCoMo-Plus.}
\begin{quote}\small\ttfamily
Earlier conversation:\\
\{CUE\_DIALOGUE\}\\[2pt]
Later message:\\
\{TRIGGER\_QUERY\}\\
Answer:
\end{quote}

\paragraph{LongMemEval.}
Sessions are selected from newest to oldest within a 12,000-character budget, then restored chronologically. A single oversized session is truncated from the end-selection pass. The manifest appends the final \texttt{Answer:} marker.
\begin{quote}\small\ttfamily
I will give you several history chats between you and a user. Please answer the question based on the relevant chat history.\\[4pt]
History Chats:\\[2pt]
\#\#\# Session \{N\}:\\
Session Date: \{SESSION\_DATE\}\\
Session Content:\\
\{ROLE\}: \{CONTENT\}\\[4pt]
Current Date: \{QUESTION\_DATE\}\\
Question: \{QUESTION\}\\
Answer:
\end{quote}

\paragraph{IFEval and FollowBench.}
IFEval retains the benchmark instruction verbatim in its dedicated evaluator; the complex-task processed prompt uses the following response suffix. FollowBench uses the instruction at the current level, while its full level-0-to-current instruction chain is retained for the official constraint evaluator.
\begin{quote}\small\ttfamily
\{IFEVAL\_PROMPT\}\\[2pt]
Response:
\end{quote}
\begin{quote}\small\ttfamily
\{INSTRUCTION\_AT\_CURRENT\_LEVEL\}\\
Answer:
\end{quote}
\section{Generalization Within Simple Factual-Conflict Tasks}
\label{app:transfer-generalization}

\subsection{Cross-Format Transfer}
\label{app:cross-format}

\begin{figure}[tbp]
  \centering

  \begin{minipage}[t]{0.48\textwidth}
    \centering
    \includegraphics[width=\linewidth]{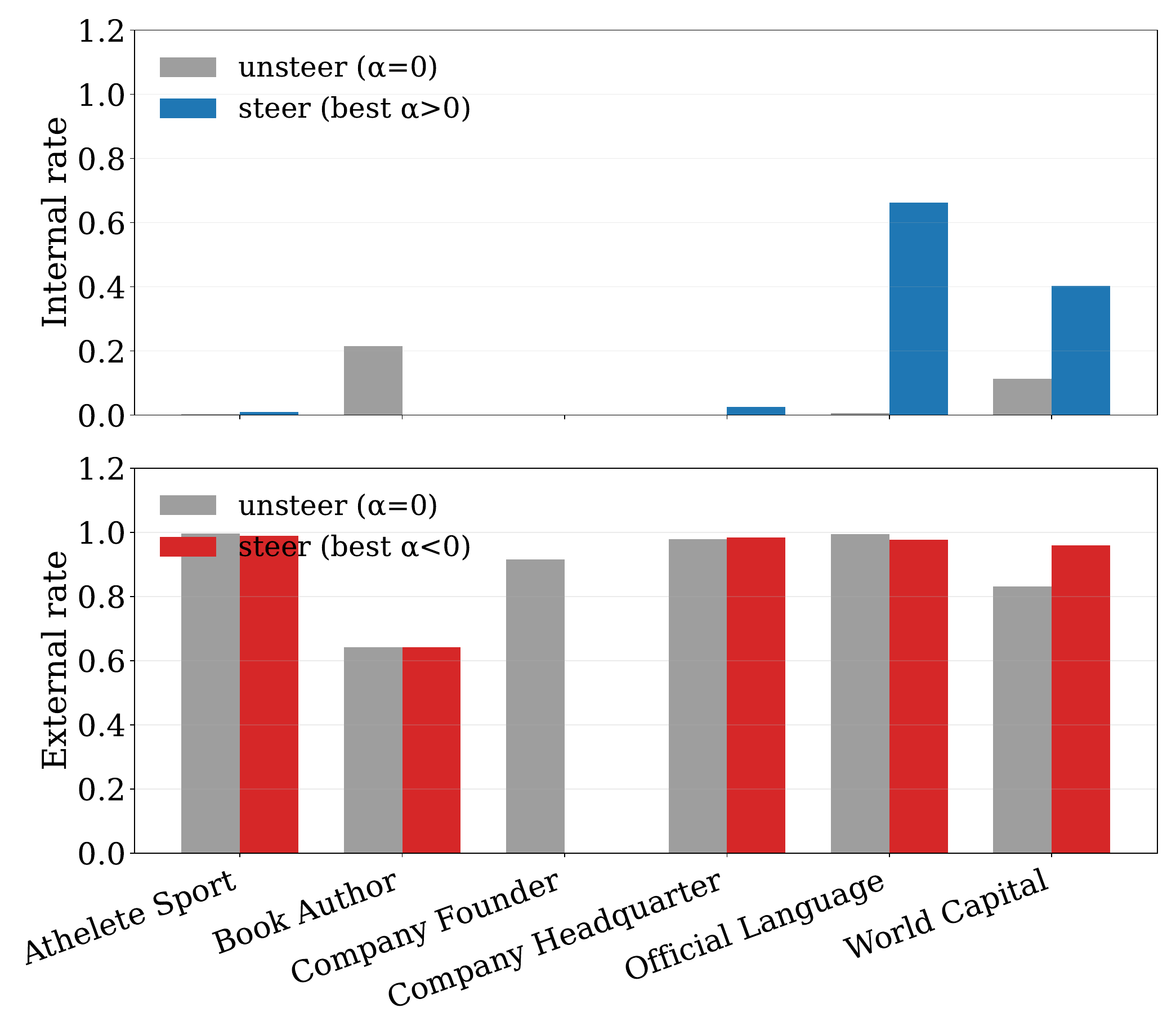}
    \small (a) Qwen3-8B
  \end{minipage}

  \vspace{0.8em}

  \begin{minipage}[t]{0.48\textwidth}
    \centering
    \includegraphics[width=\linewidth]{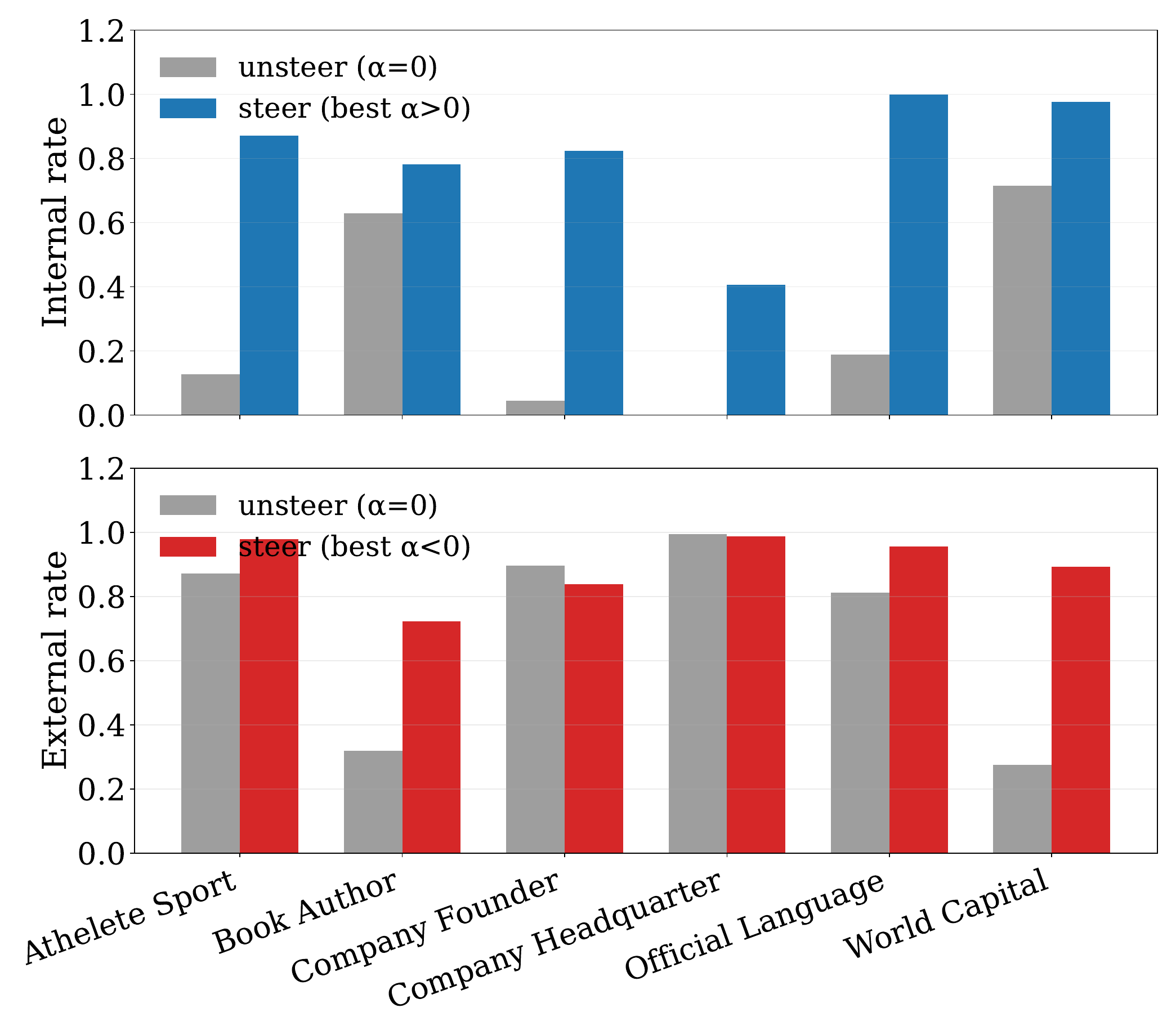}
    \small (b) Gemma-2-9B-IT
  \end{minipage}

  \vspace{0.8em}

  \begin{minipage}[t]{0.48\textwidth}
    \centering
    \includegraphics[width=\linewidth]{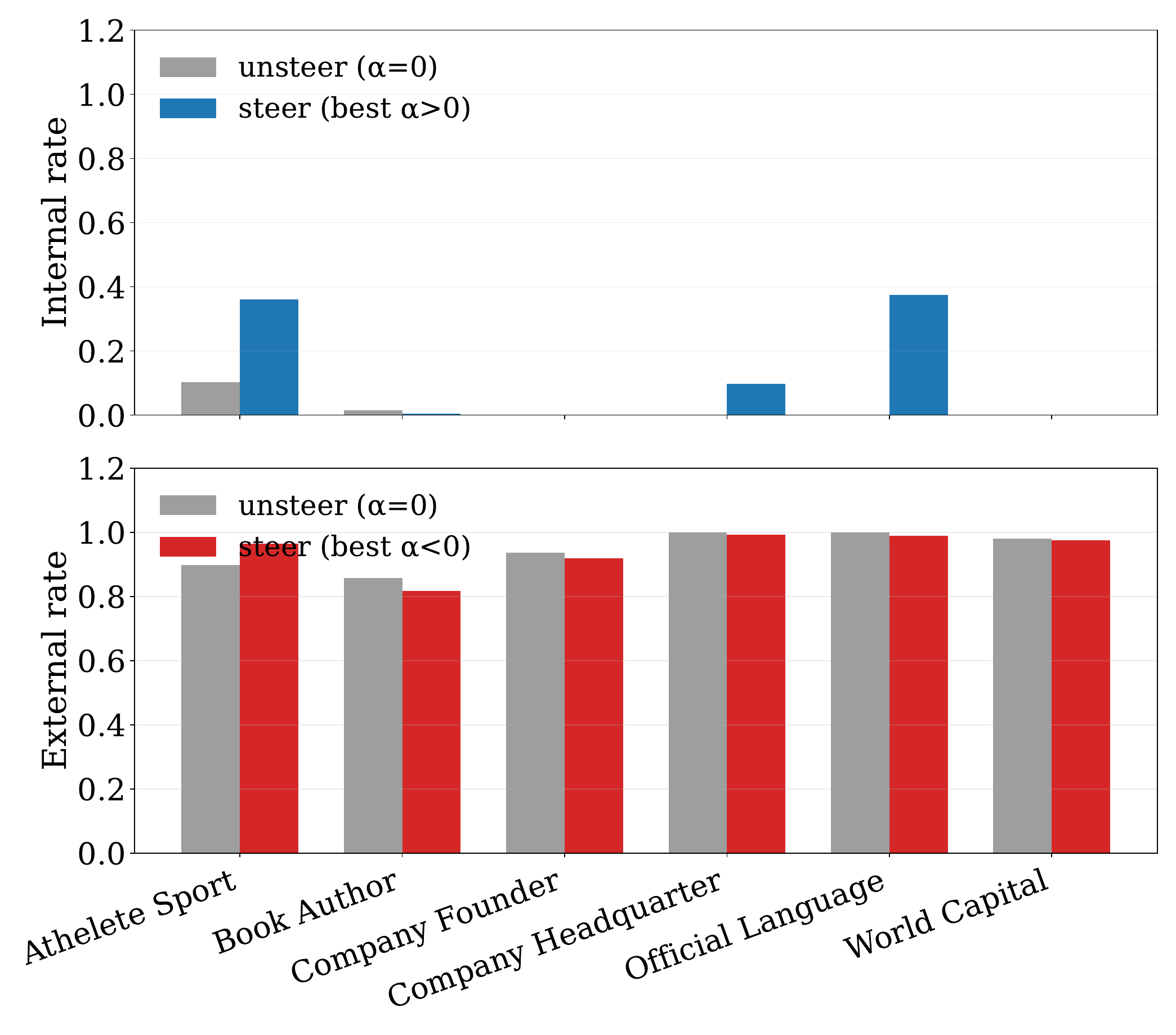}
    \small (c) Mistral-7B-Instruct-v0.3
  \end{minipage}

  \caption{Cross-format transfer on {ParaConflict} for the remaining three models. 
  Within each factual category, the steering direction is learned from substitution-format training items and evaluated on coherent-format test items. 
  Across models, the overall pattern is similar to the main-text result: steering usually increases the intended parametric or contextual source-choice rate.}
  \label{fig:cross-format-app}
\end{figure}

{ParaConflict} provides substitution and coherent conflict formats for the same questions, as described in the transfer setup. 
Figure~\ref{fig:cross-format-app} supplements the main-text results by showing cross-format transfer for the remaining three models. 
Across models, steering generally improves the selection of the intended knowledge source in most categories, indicating that the learned direction transfers across formats rather than being tied to a specific surface template.

\begin{figure*}[tbp]
  \centering
  \begin{minipage}[t]{0.7\textwidth}
    \centering
    \includegraphics[width=\linewidth]{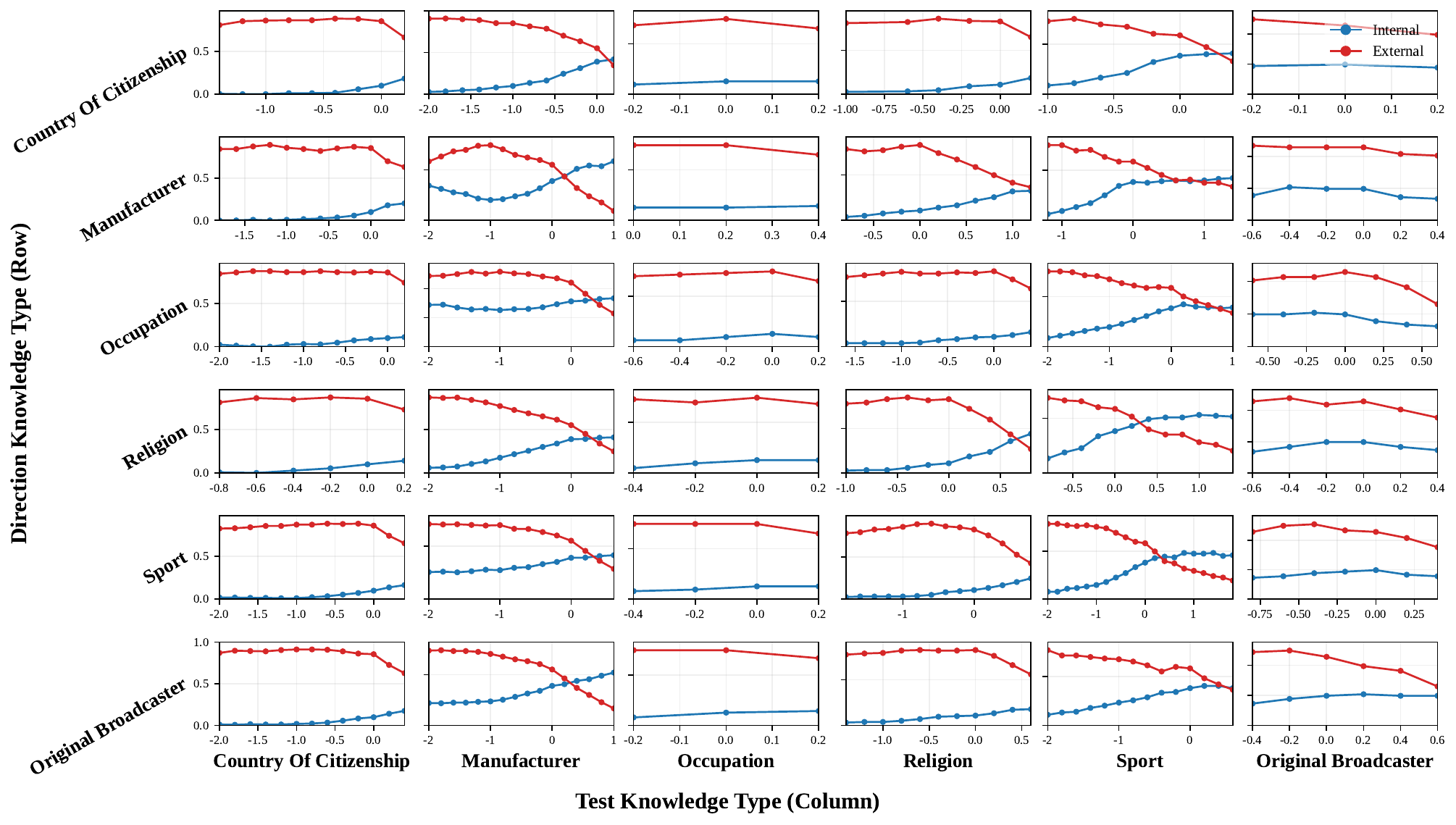}
    \small (a) Qwen3-8B
  \end{minipage}

  \vspace{0.8em}

  \begin{minipage}[t]{0.7\textwidth}
    \centering
    \includegraphics[width=\linewidth]{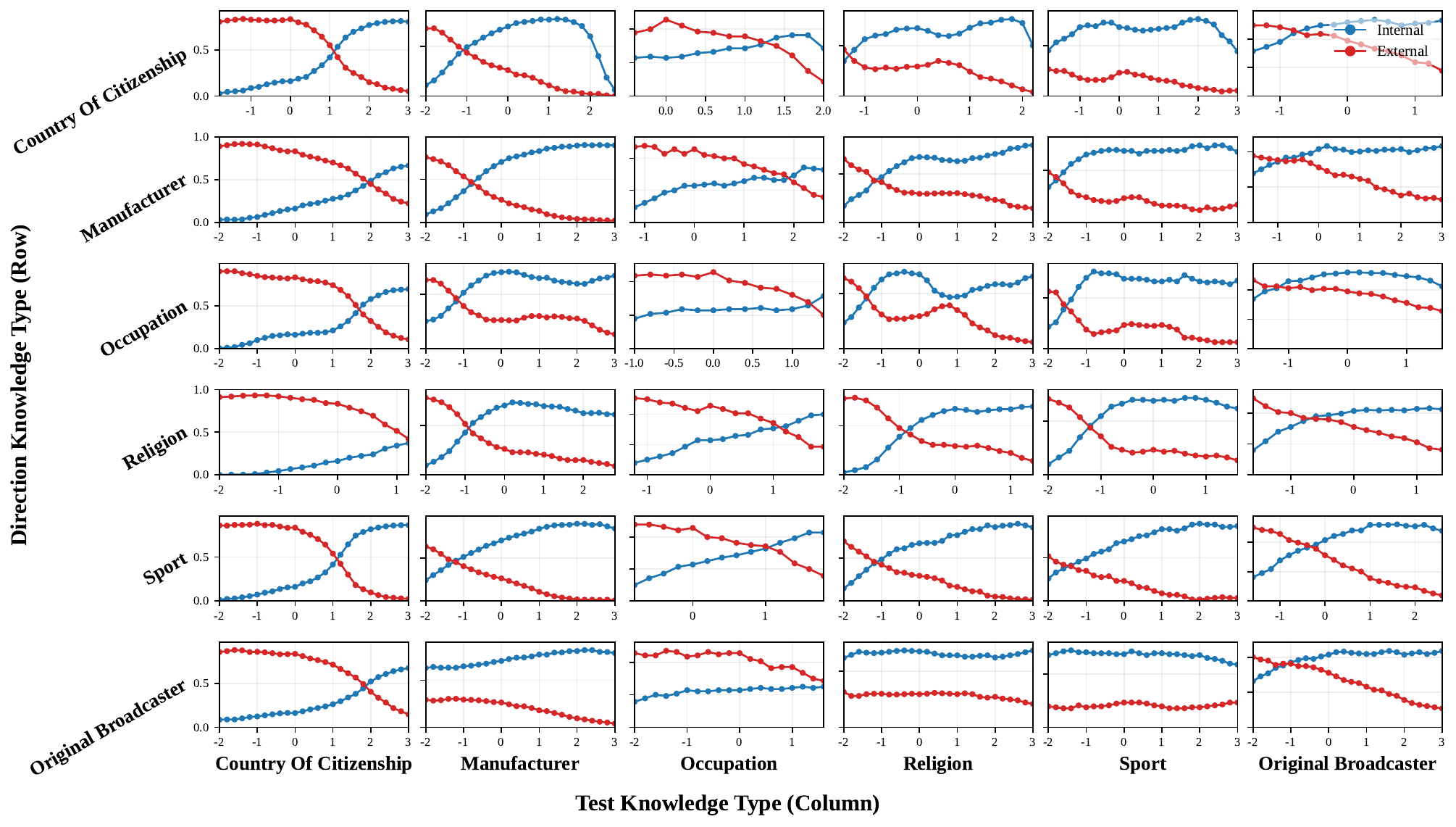}
    \small (b) Gemma-2-9B-IT
  \end{minipage}

  \vspace{0.8em}

  \begin{minipage}[t]{0.7\textwidth}
    \centering
    \includegraphics[width=\linewidth]{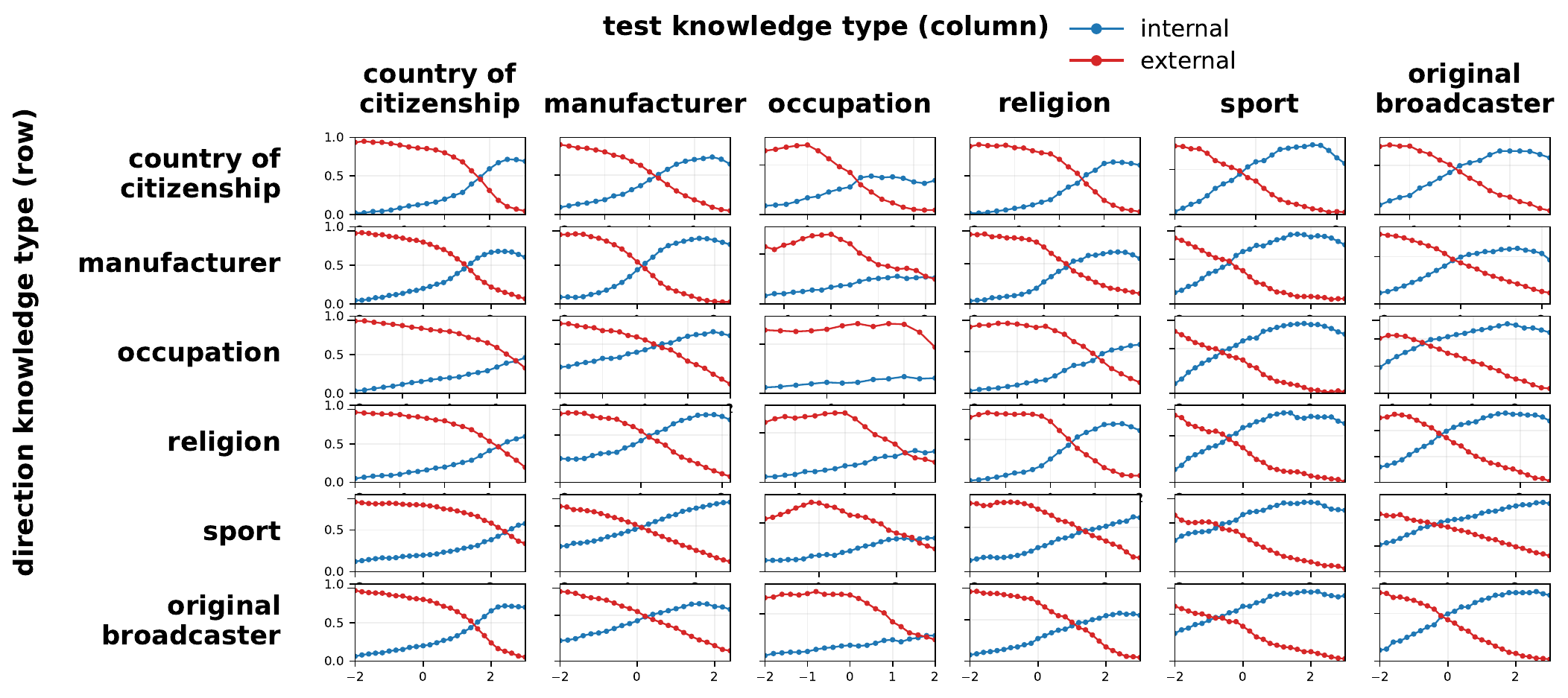}
    \small (c) Mistral-7B-Instruct-v0.3
  \end{minipage}

  \caption{Cross-knowledge-type transfer on {CounterFact} for the remaining three models. Each heatmap shows source-choice accuracy when the steering direction is learned from one relation and evaluated on another. Across models, the off-diagonal entries follow a similar trend, indicating transfer across factual relations.}
  \label{fig:cross-relation-app}
\end{figure*}
\subsection{Cross-Knowledge-Type Transfer}
\label{app:cross-knowledge-type}

Figure~\ref{fig:cross-relation-app} supplements the main-text cross-knowledge-type analysis by showing results for the remaining three models. 
Across these models, we observe a similar cross-relation transfer pattern, further supporting that the learned direction is not tied to a single factual relation.
\subsection{Cross-Dataset Transfer}
\label{app:cross-dataset}

We report additional cross-dataset transfer results for the remaining three models in Tables~\ref{tab:crossdata-gemma}, \ref{tab:crossdata-qwen}, and \ref{tab:crossdata-mistral}. 
Following the main-text setup, the steering direction and coefficients are selected only on {NQ-Swap} and then directly applied to other test datasets without target-specific tuning. 
Across models, steering generally improves both Acc$_\mathrm{par}$ and Acc$_\mathrm{ctx}$, further supporting that the learned direction captures a transferable knowledge-selection signal rather than overfitting to a single dataset.

\begin{table}[tbp]
  \centering
  \footnotesize
  \setlength{\tabcolsep}{8pt}
  \renewcommand{\arraystretch}{1.12}
  \caption{Cross-dataset transfer results on {Gemma-2-9B-it}. 
  The steering direction $\hat{v}$ and coefficients $(\alpha^{+},\alpha^{-})$ are selected on {NQ-Swap} and directly evaluated on each test data without target-specific tuning. }
  \label{tab:crossdata-gemma}
  \begin{tabular}{@{}l cc@{}}
    \toprule
    Test Data & \tabhead{Unsteered} & \tabhead{Ours} \\
    \addlinespace[2pt]
    \multicolumn{3}{@{}l}{\textit{Steer toward parametric knowledge (Acc$_\mathrm{par}$)}} \\
    \midrule
    {MacNoise} & 45.5 & \tbest{77.6} \gain{32.1} \\
    {ParaConflict} & 27.6 & \tbest{92.8} \gain{65.2} \\
    {CounterFact} & 48.0 & \tbest{76.8} \gain{28.8} \\
    {MemoTrap} & 76.1 & \tbest{78.8} \gain{2.7} \\
    \addlinespace[3pt]
    \multicolumn{3}{@{}l}{\textit{Steer toward contextual knowledge (Acc$_\mathrm{ctx}$)}} \\
    \midrule
    {MacNoise} & 25.6 & \tbest{43.0} \gain{17.4} \\
    {ParaConflict} & 70.4 & \tbest{92.8} \gain{22.4} \\
    {CounterFact} & 48.1 & \tbest{73.3} \gain{25.2} \\
    {MemoTrap} & 16.2 & \tbest{20.2} \gain{3.9} \\
    \bottomrule
  \end{tabular}
\end{table}

\begin{table}[tbp]
  \centering
  \footnotesize
  \setlength{\tabcolsep}{8pt}
  \renewcommand{\arraystretch}{1.12}
  \caption{Cross-dataset transfer results on {Qwen3-8B}. 
  The steering direction $\hat{v}$ and coefficients $(\alpha^{+},\alpha^{-})$ are selected on {NQ-Swap} and directly evaluated on each test data without target-specific tuning. }
  \label{tab:crossdata-qwen}
  \begin{tabular}{@{}l cc@{}}
    \toprule
    Test Data & \tabhead{Unsteered} & \tabhead{Ours} \\
    \addlinespace[2pt]
    \multicolumn{3}{@{}l}{\textit{Steer toward parametric knowledge (Acc$_\mathrm{par}$)}} \\
    \midrule
    {MacNoise} & 50.4 & \tbest{56.4} \gain{6.0} \\
    {ParaConflict} & 2.4 & \tbest{27.4} \gain{25.0} \\
    {CounterFact} & 26.3 & \tbest{39.8} \gain{13.5} \\
    {MemoTrap} & 83.8 & \tbest{85.8} \gain{2.0} \\
    \addlinespace[3pt]
    \multicolumn{3}{@{}l}{\textit{Steer toward contextual knowledge (Acc$_\mathrm{ctx}$)}} \\
    \midrule
    {MacNoise} & 22.6 & \tbest{33.7} \gain{11.0} \\
    {ParaConflict} & 95.8 & \tbest{96.5} \gain{0.8} \\
    {CounterFact} & 67.1 & \tbest{72.7} \gain{5.5} \\
    {MemoTrap} & 5.3 & \tbest{13.4} \gain{8.1} \\
    \bottomrule
  \end{tabular}
\end{table}

\begin{table}[tbp]
  \centering
  \footnotesize
  \setlength{\tabcolsep}{8pt}
  \renewcommand{\arraystretch}{1.12}
  \caption{Cross-dataset transfer results on {Mistral-7B-Instruct-v0.3}. 
  The steering direction $\hat{v}$ and coefficients $(\alpha^{+},\alpha^{-})$ are selected on {NQ-Swap} and directly evaluated on each test data without target-specific tuning. }
  \label{tab:crossdata-mistral}
  \begin{tabular}{@{}l cc@{}}
    \toprule
    Test Data & \tabhead{Unsteered} & \tabhead{Ours} \\
    \addlinespace[2pt]
    \multicolumn{3}{@{}l}{\textit{Steer toward parametric knowledge (Acc$_\mathrm{par}$)}} \\
    \midrule
    {MacNoise} & 51.0 & \tbest{61.6} \gain{10.6} \\
    {ParaConflict} & 3.4 & \tbest{21.4} \gain{18.0} \\
    {CounterFact} & 32.4 & \tbest{45.5} \gain{13.1} \\
    {MemoTrap} & 70.3 & \tbest{83.5} \gain{13.2} \\
    \addlinespace[3pt]
    \multicolumn{3}{@{}l}{\textit{Steer toward contextual knowledge (Acc$_\mathrm{ctx}$)}} \\
    \midrule
    {MacNoise} & 23.8 & \tbest{45.0} \gain{21.2} \\
    {ParaConflict} & 92.9 & \tbest{94.7} \gain{1.8} \\
    {CounterFact} & 65.2 & \tbest{71.9} \gain{6.7} \\
    {MemoTrap} & 27.5 & \tbest{48.4} \gain{20.9} \\
    \bottomrule
\end{tabular}
\end{table}
\end{appendices}
\end{document}